\documentclass{article}

\usepackage{amsmath, amssymb, algorithm, mathtools}
\usepackage{bbm, graphicx, float, pifont,multirow}
\usepackage[printonlyused]{acronym}

\usepackage[squaren,Gray]{SIunits}
\usepackage{pgf,tikz,paralist}
\usetikzlibrary{bayesnet,arrows,automata,positioning}
\usepackage[normalem]{ulem}

\usepackage{setspace}
\usepackage[left=3.5cm,right=3.5cm,top=3cm,bottom=4cm]{geometry}

\usepackage[printonlyused]{acronym}
\pdfoutput=1

\newcommand{\cf}{c.f.~}
\newcommand{\eg}{e.g.,~}
\newcommand{\ie}{i.e.,~}

\newcommand{\apriori}{\textit{a priori}}

\newcommand{\SNR}{\text{SNR}}

\newcommand{\iid}{iid.~}

\newcommand{\rreal}{\mathbb{R}}

\newcommand{\diag}[1]{\text{diag}{\left(#1\right)\,}}

\newcommand{\vs}[1]{\boldsymbol{\mathrm{#1}}}

\newcommand{\paramIllum}{\beta_\text{IP}}

\newcommand{\algref}[1]{Alg.~\ref{#1}}
\newcommand{\eeqref}[1]{Eq.~(\ref{#1})}
\newcommand{\figref}[1]{Fig.~\ref{#1}}

\newcommand{\secref}[1]{Section~\ref{#1}}

\newcommand{\tabref}[1]{Tab.~\ref{#1}}

\newcommand{\expf}[1]{\exp\{#1\}}

\newcommand{\norm}[1]{\|#1\|}

\newcommand{\RMSE}[1]{\epsilon_{#1}}

\newcommand{\numFeats}{K}

\newcommand{\prodN}[0]{\prod_{n=1}^N}

\newcommand{\prodK}[0]{\prod_{k=1}^K}
\newcommand{\prodKstar}[0]{\prod_{k=1}^{K^\star}}

\newcommand{\sumN}[0]{\sum_{n=1}^N}

\newcommand{\sumD}[0]{\sum_{d=1}^D}
\newcommand{\sumDt}[0]{\sum_{d'=1}^D}
\newcommand{\sumK}[0]{\sum_{k=1}^K}

\newcommand{\sumKt}[0]{\sum_{k'=1}^K}
\newcommand{\sumKtt}[0]{\sum_{k''=1}^K}
\newcommand{\sumKtwok}[0]{\sum_{\substack{k'=1\\k' \neq k}}^K}

\newcommand{\pdf}[0]{p}
\newcommand{\pms}[0]{P}

\newcommand{\betapdf}[2]{ \text{Beta}_{#1}\!\left(#2\right)\,}

\newcommand{\dirpdf}[2]{ \text{Dir}_{#1}\!\left(#2\right)\,}

\newcommand{\gampdf}[2]{ \text{Ga}_{#1}\!\left(#2\right)\,}
\newcommand{\gausspdf}[3]{\mathcal{N}_{#1}\!\left(#2,#3\right)\,}
\newcommand{\invgampdf}[2]{ \text{IGa}_{#1}\!\left(#2\right)\,}

\newcommand{\poissonpdf}[2]{ \text{Poisson}_{#1}\!\left(#2\right)\,}
\newcommand{\truncgausspdf}[2]{\mathcal{TN}_{#1}\!\left(#2\right)\,}
\newcommand{\uniformpdf}[2]{\mathcal{U}_{#1}\!\left(#2\right)\,}

\newcommand{\trans}[1]{{#1}^{\text{T}}}
\newcommand{\inv}[1]{{#1}^{-1}}

\newcommand{\arccosfun}[1]{\text{arccos}\!\left(#1\right)\,}
\newcommand{\logfun}[1]{\text{log}\!\left(#1\right)\,}

\newcommand{\betafun}[1]{\text{B}\!\left(#1\right)\,}
\newcommand{\indicatorfun}[1]{\vs 1\!\left(#1\right)\,}

\newcommand{\positivefun}[1]{H\!\left(#1\right)\,}

\newcommand{\Amat}{ {\vs A} }
\newcommand{\AFmat}{ {\vs A^\star} }

\newcommand{\Imat}{ {\vs I} }
\newcommand{\Fmat}{ {\vs F} }

\newcommand{\Smat}{ {\vs S} }

\newcommand{\Wmat}{ {\vs W} } 
\newcommand{\Zmat}{ {\vs Z} }

\newcommand{\Zclass}{ {Z_{[\AFmat]}}}

\newcommand{\SID}{ \text{SID} }

\newcommand{\sigmaZ}{ \sigma_{z} }

\newcommand{\hyphypAalphaA}{ {h^{(1)}_{\alpha_a}} } % hyper hyper on alpha_a
\newcommand{\hyphypAalphaB}{ {h^{(2)}_{\alpha_a}} }
\newcommand{\hyphypAbetaA}{ {h^{(1)}_{\beta_a}} } % hyper hyper on beta_a
\newcommand{\hyphypAbetaB}{ {h^{(2)}_{\beta_a}} }

\newcommand{\hyphypSigmaAlphaA}{ {h^{(1)}_{\alpha_\sigma}} }
\newcommand{\hyphypSigmaAlphaB}{ {h^{(2)}_{\alpha_\sigma}} }
\newcommand{\hyphypSigmaBetaA}{ {h^{(1)}_{\beta_\sigma}} }
\newcommand{\hyphypSigmaBetaB}{ {h^{(2)}_{\beta_\sigma}} }

\newcommand{\hyperAalpha}{ {\alpha_a} }
\newcommand{\hyperAbeta}{ {\beta_a} }

\newcommand{\hypertauA}{ {\alpha_\sigma} }
\newcommand{\hypertauB}{ {\beta_\sigma} }

\newcommand{\hyperWgamma}{ {\gamma_w} }

\newcommand{\av}[0]{\vs a}
\newcommand{\fv}[0]{\vs f}
\newcommand{\sv}[0]{\vs s}

\newcommand{\wv}[0]{\vs w}

\newcommand{\zv}[0]{\vs z}

\acrodef{AIC}{Akaike Information Criterion}
\acrodef{BLFM}{Bayesian Latent Feature Model}
\acrodef{BLU}{Bayesian Linear Unmixing}
\acrodef{BNMF}{Bayesian Nonnegative Matrix Factorization}
\acrodef{BNU}{Bayesian Nonparametric Unmixing}
\acrodef{BSS}{Blind Source Separation}
\acrodef{ELM}{Eigenvalue Likelihood Maximization}
\acrodef{IBP}{Indian Buffet Process}	
\acrodef{FL}{Feature Learning}
\acrodef{hideNN}{Denoised Hyperspectral Intrinsic Dimensionality Estimation With Nearest-Neighbor Distance Ratios}
\acrodef{HSI}{Hyperspectral Imaging}
\acrodef{HSU}{Hyperspectral Unmixing}
\acrodef{HySime}{Hyperspectral Signal Subspace Identification by Minimum Error}
\acrodef{ICE}{Iterated Constrained Endmembers}
\acrodef{MAP}{\textit{maximum-a-posteriori}}
\acrodef{MCMC}{Markov Chain Monte Carlo}
\acrodef{MDL}{Minimum Description Length}
\acrodef{MMSE}{Minimum Mean Squared Error}
\acrodef{MVES}{Minimum-Volume Enclosing Simplex}
\acrodef{MVT}{Minimum Volume Transform}
\acrodef{NMF}{Non-negative Matrix Factorization}
\acrodef{PCA}{Principal Component Analysis}
\acrodef{pdf}{probability density function}
\acrodef{pmf}{probability mass function}
\acrodef{PPI}{Pixel Purity Index}
\acrodef{PT}{Parallel Tempering}
\acrodef{RMSE}{Root Mean Squared Error}
\acrodef{SNR}{Signal-to-Noise Ratio}
\acrodef{SPICE}{Sparsity-Promoting ICE}
\acrodef{VCA}{Vertex Component Analysis}
\acrodef{VD}{Virtual Dimensionality}

\begin{document}

\title{Bayesian Nonparametric Unmixing\\ of Hyperspectral Images}

\author{J{\"u}rgen~Hahn\footnote{jhahn@spg.tu-darmstadt.de} \quad 
	Abdelhak~M.~Zoubir\footnote{zoubir@spg.tu-darmstadt.de}\\
	Signal Processing Group\\ Institute of Telecommunications, Technische Universit{\"a}t Darmstadt\\ Merckstra{\ss}e 25, 64283 Darmstadt}
\date{}

\maketitle

\begin{abstract}
Hyperspectral imaging is an important tool in remote sensing, allowing for accurate analysis of vast areas.
Due to a low spatial resolution, a pixel of a hyperspectral image rarely represents a single material, but rather a mixture of different spectra.
\ac{HSU} aims at estimating the pure spectra present in the scene of interest, referred to as endmembers, and their fractions in each pixel, referred to as abundances.
Today, many \ac{HSU} algorithms have been proposed, based either on a geometrical or statistical model. 
While most methods assume that the number of endmembers present in the scene is known, there is only little work about estimating this number from the observed data.
In this work, we propose a Bayesian nonparametric framework that jointly estimates the number of endmembers, the endmembers itself, and their abundances, by making use of the Indian Buffet Process as a prior for the endmembers.
Simulation results and experiments on real data demonstrate the effectiveness of the proposed algorithm, yielding results 
comparable with state-of-the-art methods while being able to reliably infer the number of endmembers.
In scenarios with strong noise, where other algorithms provide only poor results, the proposed approach tends to overestimate the number of endmembers slightly. 
The additional endmembers, however, often simply represent noisy replicas of present endmembers and could easily be merged in a post-processing step.
\end{abstract}

\acresetall

\section{Introduction}
In \ac{HSI}, the reflected light of the scene of interest is captured by mapping the spectral range of light to a finite number of continuous bands. Hence, the captured hyperspectral image represents the reflected spectrum at any location in the scene. 
Since each material possesses a characteristic spectrum, also referred to as signature, scene analysis based on \ac{HSI} becomes fairly easy \cite{Shaw2002}. Today, \ac{HSI} is mainly used in airborne and spaceborn remote sensing \cite{Green1998,Pearlman2000}, e.g. for agriculture, urban mapping, and security applications \cite{Schowengerdt2007}. 

Especially in remote sensing, the captured images often suffer from a low spatial resolution. 
Thus, an element of the image, a pixel, represents a mixture of different materials. 
However, most post-processing algorithms that are used for data analysis assume pure pixels, \ie each pixel is assumed to represent a single material.
This is often the case, \eg in classification \cite{Damodaran2015, Hahn2014, Hahn2014a}.
For this reason, \ac{HSU} is an important task for the analysis of hyperspectral images, revealing the endmembers and their abundances present in the scene of interest. 
The endmembers can be understood as the raw materials occurring in the scene, while the abundances describe the fractions of which the endmembers are present in each pixel. 

In the past, various algorithms have been developed to solve the problem of \ac{HSU}. An excellent overview is given in \cite{Bioucas-Dias2012}. 
Many methods split the \ac{HSU} problem into two separate tasks: (i) endmember extraction and (ii) abundance estimation, \eg \ac{PPI} \cite{Boardman1993}, N-FIND-R \cite{Winter1999}, and \ac{VCA} \cite{Nascimento2005}. 
Bayesian methods provide means for performing these tasks jointly. In \cite{Arngren2010}, a Bayesian framework is presented for jointly inferring the endmembers and abundances. Different priors for the endmembers are investigated, motivated by regularization terms of existing nonprobabilistic unmixing methods. 
A different Bayesian framework is proposed in \cite{Dobigeon2009}, in which the fact is exploited that the abundances lie in a subspace.
Other examples of algorithms that aim at solving this task simultaneously are \ac{ICE} \cite{Berman2004}, \ac{MVT} \cite{Craig1994}, \ac{MVES} \cite{Chan2009} and \ac{NMF} \cite{Lee2001,Pauca2006}.
While these models assume a linear relationship between endmembers and abundances, recent work provides methods for nonlinear models \cite{Halimi2011, Altmann2013, Dobigeon2014, Altmann2015}. 
Further, semi-supervised approaches have been explored, where the endmembers in the scene are selected from a dictionary instead of being learned \cite{Dobigeon2008, Themelis2012}.

Though there already exist many methods and algorithms that aim at solving the unmixing problem, most of them assume that the number of materials in the scene is known \apriori. However, this assumption is hardly fulfilled in practice. If the number is set incorrectly and differs from the true number, most methods will try to fit the observed data into an incorrect model, which may yield poor results. Especially an underestimate of the number is critical, as then endmembers present in the scene are simply not extracted and remain undiscovered. 

The few work that aims at estimating the number of endmembers is mainly based on subspace methods \cite{Bioucas-Dias2012}. 
In \cite{Chang2004}, it is reported that classical methods for model-order selection, such as \ac{AIC} \cite{Akaike1974} and \ac{MDL} \cite{Schwarz1978}, do not work well in the context of \ac{HSU} due to their assumptions on the noise.
Advanced methods for model selection such as those based on the bootstrap do not require assumptions on the noise \cite{Zoubir1998, Zoubir2004}. However, these methods are computationally intensive. 

Thus, new algorithms for \ac{HSI} have been developed. The probably most prominent method is \ac{VD} \cite{Chang2004}, where information theoretic criteria are utilized together with a Neyman-Pearson test to detect the number of endmembers.
\ac{HySime} \cite{Bioucas-Dias2008} aims at estimating the signal subspace by minimizing the projection errors of the signal and noise subspace. The dimensionality of the signal subspace can then be understood as an estimate of the number of endmembers.
In \cite{Luo2013}, \ac{ELM} is presented, which infers the number of endmembers by means of a comparison between the correlation and covariance matrix of the spectra. 
A \ac{SPICE} is proposed in \cite{Zare2007}, extending the \ac{ICE} algorithm by placing a sparsity-promoting prior on the abundances.
A thresholding scheme is then applied to the abundances and endmembers are pruned if not present in the scene. 
In \cite{Heylen2013}, the authors explain a geometric approach based on nearest neighbors for dimensionality estimation of a manifold. Due to the sensitivity to noise, they further introduce a denoised version of the algorithm, the \ac{hideNN}.

We argue that \ac{HSU} can also be considered as a feature learning problem, where the features represent the endmembers and the coefficients of the features the abundances. 
A \ac{BLFM} has been proposed in \cite{Schmidt2009a}, which has been adjusted in \cite{Arngren2010} for \ac{HSU}. 
A nonparametric version of \ac{BLFM} is developed in \cite{Ghahramani2005,Knowles2011}, which allows to infer the number of features from the observations. 
As explained above, inferring the number of latent endmembers is a highly desirable property of any \ac{HSU} algorithm, giving rise to a fully-automated approach for \ac{HSU}.
Therefore, we follow the approach in \cite{Knowles2011} and extend the Bayesian framework in \cite{Arngren2010} by placing an \ac{IBP} prior on the activations of the endmembers, resulting in a Bayesian nonparametric model \cite{Gershman2012}. 
The \ac{IBP} describes an infinite feature model, while the number of features drawn from this process is always finite. 
Thus, the \ac{IBP} provides means for inferring the number of present endmembers in the scene.
Our proposed algorithm, \ac{BNU}, allows for the joint inference of the endmembers, their abundances, and also the number of endmembers, in contrast to most existing \ac{HSU} algorithms. 

This work is structured as follows. In \secref{sec::ibp}, we shortly revisit the \ac{IBP}.
Based on \cite{Dobigeon2009,Arngren2010}, we provide a Bayesian nonparametric model for \ac{HSU} in \secref{sec::model} and show in \secref{sec::inference} how to perform inference in this model. \secref{sec::results} provides results on synthetic as well as real data. 
In \secref{sec::discussion}, we comment on the results and provide insights for future directions. Finally, conclusions are drawn.

\section{Indian Buffet Process}
\label{sec::ibp}
The \acf{IBP} \cite{Ghahramani2005} describes a model for sampling a sparse binary feature matrix, assuming an infinite number of features. 
In this section, we focus on the main results of the two-parameter generalization \cite{Ghahramani2007}. 
The full derivation of the \ac{IBP} is given in \cite{Ghahramani2005,Griffiths2011}. 
In the following, variables with a star ($^\star$) belong to the finite feature model, the bracket $[\cdot]$ denotes the class of feature representations, and variables without a star are either independent of the feature model or belong to the infinite feature model.

For a finite number of features $K^\star$, the sums of the rows of the feature activation matrix $\AFmat \in \{0,1\}^{K^\star \times D}$ follow \iid Binomial distributions, with $D$ denoting the dimension of the features, \ie the number of bands. 
Placing a Beta prior with hyperparameters $\frac{\hyperAalpha \hyperAbeta}{K^\star}$ and $\hyperAbeta$ over the parameter $\theta_a$ of the Bernoulli distribution and marginalizing over $\theta_a$ yields a Beta-Binomial distribution \cite{Ghahramani2005, Ghahramani2007},
\begin{align}
 \begin{split}
 \pms(\AFmat  \,|\, \hyperAalpha,\hyperAbeta) & = \prodKstar \int_0^1 \! \pms(\av_k^\star \,|\, \theta_a) \pdf(\theta_a \,|\,  \frac{\hyperAalpha \hyperAbeta}{K^\star}, \hyperAbeta) \, \mathrm{d} \theta_a
		    \\ & = \prodKstar \frac{\betafun{ m_k^\star + \frac{\hyperAalpha \hyperAbeta}{K^\star}, D - m_k^\star + \hyperAbeta} }{ \betafun{ \frac{\hyperAalpha \hyperAbeta}{K^\star}, \hyperAbeta}},
 \end{split}		    
 \label{eq::ibp::finitemodel}
\end{align}
where $m_k^\star$ counts the number of ones in the $k$th row of $\AFmat$ and $\betafun{a,b}$ is the Beta function with parameters $a$ and $b$.

Eventually, we aim at sampling feature representations.
The distribution in \eeqref{eq::ibp::finitemodel}, however, describes the probability of binary matrices, where different realizations may describe an equivalent feature representation. 
In particular, permutations of the rows of $\AFmat$ belong to the same representation. 
According to \cite{Ghahramani2005}, the probability of a feature representation,  $[\AFmat]$, is given by
\begin{align*}
  \pms([\Amat^\star]\,|\,\hyperAalpha,\hyperAbeta) &= \frac{1}{\Zclass} \pms(\Amat^\star\,|\,\hyperAalpha,\hyperAbeta),
\end{align*}
with normalization $\Zclass$,
\begin{align*}
  \Zclass = \begin{pmatrix} K^\star \\ \prod_{\vs h \in \{0,1\}^D } K_h \end{pmatrix},
\end{align*}
where $K_h$ denotes the number of occurrences of the binary vector $\vs h \in \{0,1\}^D$.
Since we are interested in sampling from an infinite number of features, we consider the limit for \mbox{$K^\star \rightarrow \infty$} \cite{Ghahramani2005, Doshi-Velez2009}, 
\begin{align}
 \begin{split}
 \pms([\Amat]\,|\,\hyperAalpha,\hyperAbeta) &= \lim_{K^\star \rightarrow \infty} \pms([\Amat^\star]\,|\,\hyperAalpha,\hyperAbeta)
 \\ & = \frac{(\hyperAalpha \hyperAbeta)^{K}}{\prod_{\vs h \in \{0,1\}^D \backslash \vs 0} K_h!} \expf{-\bar{K}} 
 \\ & \quad  \times \prodK \betafun{m_k, D - m_k + \hyperAbeta} \!,      
 \end{split}
 \label{eq::ibp::infinitemodel}
\end{align}
where $K$ denotes the number of active rows and \mbox{$\bar{K} = \hyperAalpha \sumD \frac{\hyperAbeta}{\hyperAbeta + d -1}$} is the expected number of active rows of $\Amat$ . Both hyperparameters, $\hyperAalpha$ and $\hyperAbeta$, increase the probability of an active entry in $\Amat$. 
At the same time, the first hyperparameter, $\hyperAalpha$, controls the expected number of features, resulting in sparse realizations.
The second hyperparameter, $\hyperAbeta$, permits to decouple the feature generation from the sparsity, allowing for dense as well as sparse realizations of $\Amat$ \cite{Ghahramani2007,Griffiths2011}. 
We want to emphasize that, consequently, the \ac{IBP} promotes, but does not enforce, sparse matrices and, hence, allows for dense rows realizations of $\Amat$.

\subsection{Sampling from the Indian Buffet Process}
Sampling from \eeqref{eq::ibp::infinitemodel} is easily performed using a Gibbs sampler \cite{Ghahramani2007} with a subsequent ordering of the activations.
From the finite model in \eeqref{eq::ibp::finitemodel}, the conditional for sampling an element of the finite feature activation matrix, $a^\star_{k,d},$ with $k=1,\ldots,K^\star$ and $d= 1,\ldots, D$, can be derived as \cite{Ghahramani2007}
\begin{align*}
  \pms(a^\star_{k,d}  = 1 \,|\, \vs a^\star_{k \backslash d} ) &= \frac{ m^\star_{k\backslash d} + \frac{\hyperAalpha \hyperAbeta}{K^\star}}{D + \frac{\hyperAalpha \hyperAbeta}{K^\star} + \hyperAbeta -1},
\end{align*}
where $\av^\star_{k\backslash d}$ is the $k$th row of $\AFmat$ without $a^\star_{k,d}$ and $m^\star_{k\backslash d}$ is the sum over the elements of $\av^\star_{k\backslash d}$.
Considering the limit for $K^\star \rightarrow \infty$ results in \cite{Ghahramani2007}
\begin{align*}
  \pms(a_{k,d} = 1\,|\, \vs a_{k \backslash d} ) &= \frac{ m_{k\backslash d}}{D + \hyperAbeta -1 }.
\end{align*}
Note that there is a certain probability that every object, \ie every hyperspectral band, has been generated by a feature that has not been inferred yet.
Assuming exchangeability, the ordering of the variables $a_{k,d}$ becomes irrelevant \cite{Ghahramani2005}.
Thus, the probability of activating $K^+$ new features for the $d$th band is 
\begin{align}
   \pms(K^+\,|\, -) \sim \poissonpdf{K^+}{\frac{\hyperAalpha \hyperAbeta}{\hyperAbeta + D - 1}}\!\!,
   \label{eq::ibp::newfeat}
\end{align}
where the bar symbol ($-$) refers to all random variables except $K^+$.

In summary, sampling works as follows.
We set $a_{k,d}$ to one with probability $\frac{ m_{d \backslash k}}{D + \hyperAbeta -1}$. 
With probability $\pms(K^+\,|\, -)$, we add $K^+$ elements to the $d$th column. 
After having iterated over all active rows, a proposal is made to remove all columns that contain zero entries only, resulting in a sample of a binary matrix with $K$ active rows, \ie $K$ features.

Note that the samples generated by means of this algorithm need to be ordered if we want to sample from feature class representations \cite{Griffiths2011}.

\subsection{Sampling the hyperparameters $\hyperAalpha$ and $\hyperAbeta$ }
\label{sec::ibp::sampling::hyper}
The hyperparameters $\hyperAalpha$ and $\hyperAbeta$ can be considered as variables that are Gamma distributed with hyperparameters $\hyphypAalphaA, \hyphypAalphaB$ and $\hyphypAbetaA, \hyphypAbetaB$, respectively \cite{Knowles2011}.
Thus, the conditional of $\hyperAalpha$ is given by 
\begin{align*}
 \pdf(\hyperAalpha \,|\, -) = \gampdf{\hyperAalpha}{K + \hyphypAalphaA, \sumD \frac{\hyperAbeta}{\hyperAbeta + d -1} + \hyphypAalphaB}\!.
\end{align*}
In order to sample $\hyperAbeta$, a Metropolis step is used with hyperprior $\pdf(\hyperAbeta) = \gampdf{\hyperAbeta}{\hyphypAbetaA, \hyphypAbetaB}$ as proposal distribution. 
The acceptance ratio $r_\hyperAbeta$ is then given by
\begin{align*}
 r_\hyperAbeta = \frac{\pdf(\hyperAbeta'\,|\,-)}{\pdf(\hyperAbeta\,|\,-)} = \frac{\pms(\Amat\,|\,\hyperAalpha,\hyperAbeta')}{\pms(\Amat\,|\,\hyperAalpha,\hyperAbeta)},
\end{align*}
where $\hyperAbeta'$ denotes the proposed value.

\section{Bayesian Nonparametric Unmixing Model}
\label{sec::model}
Given $N$ observed spectra, $\zv_n \in \rreal^{1 \times D}, n=1,\ldots,N$, of a hyperspectral image with $N$ pixels, we consider a linear unmixing problem with additive noise $\vs e_n \in \rreal^{1 \times D}$, \ie
\begin{align*}
 \zv_n = \sv_n \Fmat + \vs e_n,
\end{align*}
where $\Fmat \in \rreal_+^{K \times D}$ are the endmembers and $\sv_n \in [0, 1]^{1\times K}$ the corresponding abundances. 
The set of positive real numbers including zero is denoted by $\rreal_+$. 
The abundances are required to fulfill the additivity constraint, \ie $\sumK s_{n,k}=1$, and the positivity constraint, \ie $s_{n,k} \geq 0$, with $n=1,\ldots,N$ and $k=1,\ldots,K$, as they represent the fractions of which the endmembers occur in each pixel. 
The noise term $\vs e_n$ is assumed to be \iid Gaussian distributed, \ie $\pdf(\vs e_n\,|\,\sigmaZ) = \gausspdf{e_n}{0}{\sigmaZ^2 \Imat}$ with variance $\sigmaZ^2$.
Though this model does not capture correlated noise, it has been widely used in unmixing methods, \eg \cite{Dobigeon2009, Chang1998}.

Further, we assume that the endmember matrix $\Fmat$ is a (finite) realization of a random process modeling an infinite number of endmembers, which is described by the \ac{IBP}. 
Since the \ac{IBP} is able to present binary values only, we use an element-wise multiplication to introduce weights on the endmembers by means of the weight matrix $\Wmat \in \rreal_+^{K \times D}$ as suggested in \cite{Knowles2011}, \ie
\begin{align*}
 \Fmat = \Amat \odot \Wmat,
\end{align*}
where $\odot$ represents the element-wise matrix multiplication, the Hadamard-product.
Note that the sparsity assumption implied by the \ac{IBP} does not necessarily lead to sparse realizations of $\Fmat$.
This assumption basically states that the underlying process models an infinite but sparse matrix which is eventually of finite size and can, therefore, be stored in memory. 
The sparsity of the realization of the activation matrix, $\Amat$, is controlled by means of the hyperparameters $\hyperAalpha$ and $\hyperAbeta$ as well as by the observation likelihood. 

In the following, we detail the components of the proposed hierarchical Bayesian nonparametric model for spectral unmixing.

\begin{figure}
 \centering
 \includegraphics[width=.45\textwidth]{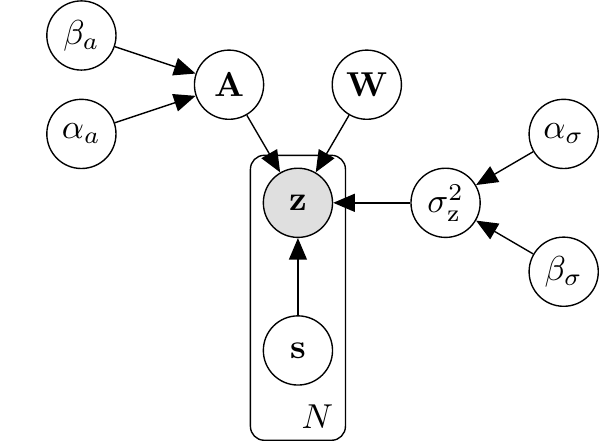}
 \caption{Graphical model of the hierarchical \acf{BNU} model. Only the spectra $\zv_n, n=1,\ldots,N$, are observed, the other variables are latent and need to be inferred. }
 \label{fig::model::graphicalmodel}
\end{figure}

\subsection{Likelihood}
We assume that the observations are conditionally independently distributed and corrupted by additive Gaussian noise. Hence, the likelihood is given as
\begin{align}
 \pdf(\Zmat\,|\, \Wmat, \Amat, \Smat, \sigmaZ^2) = \prodN \gausspdf{\zv_n}{\sv_n \left( \Amat \odot \Wmat\right)}{\sigmaZ^2 \Imat}\!\!,
 \label{eq::model::likelihood}
\end{align}
with $\Zmat = \trans{\begin{bmatrix} \trans{\zv_1} & \ldots & \trans{\zv_N} \end{bmatrix}}$ and $\Smat = \trans{\begin{bmatrix} \trans{\sv_1} & \ldots & \trans{\sv_N} \end{bmatrix}}$.

In practice, the pixels of the image may suffer also from different lighting conditions.
Deriving suitable models is challenging and, even if a suitable model was utilized, inference would probably be less efficient. 
In \secref{sec::results}, we simulate varying light conditions by means of multiplicative noise and investigate the effect on the estimates.

\subsection{Prior for the noise variance $\sigmaZ^2$}
Since $\sigmaZ^2$ is the variance of Gaussian distributed noise, a conjugate prior for $\sigmaZ^2$ is the Inverse-Gamma distribution with parameters $\hypertauA$ and $\hypertauB$, 
\begin{align*}
 \pdf(\sigmaZ^2 \,|\, \hypertauA, \hypertauB) = \invgampdf{\sigmaZ^2}{\hypertauA, \hypertauB}\!\!.
\end{align*}
Further, we assume that the hyperparameters $\hypertauA$ and $\hypertauB$ follow Gamma distributions, \ie $\hypertauA \sim \gampdf{\hypertauA}{\hyphypSigmaAlphaA, \hyphypSigmaAlphaB}$ and $\hypertauB \sim \gampdf{\hypertauB}{\hyphypSigmaBetaA, \hyphypSigmaBetaB}$, respectively.

\subsection{Prior for the abundances $\Smat$}
The prior on the fractional abundances must fulfill the additivity constraint, $\sumK s_{k,n} = 1$, and the positivity constraint, $s_{k,n} > 0$ with $k=1,\ldots,K$ and $n=1,\ldots,N$. 
The constraints can be interpreted as $\sv_{n}$ representing a probability distribution over the presence of the endmembers, giving rise to model the rows of $\Smat$ with Dirichlet distributions. 
The abundances of the pixels are assumed to be \iid yielding
\begin{align*}
    \pdf(\Smat) & = \prodN \pdf(\sv_n)
                           \\ & = \prod_{n=1}^N \dirpdf{s_{n,1}, \ldots, s_{n,K}}{\alpha_{s,1}, \ldots, \alpha_{s,K}}\!\!.    
\end{align*}
We set the hyperparameters $\alpha_{s,k} = 1$ for $k = 1,\ldots,K$, making the prior uniform under the additivity and positivity constraints.
This does not impose any preferences on the different endmembers and allows for efficient sampling as we explain in \secref{sec::inference}. 

\subsection{Prior for the endmember weights $\Wmat$ and activations  $\Amat$}
We choose the distance prior for the endmember weights $\Wmat$ with hyperparameter $\hyperWgamma$ as in \cite{Arngren2010} which can be interpreted as a probabilistic version of the volume regularization based on the Euclidean distance proposed in \cite{Berman2004}.
Since endmember spectra are positive valued, we use the following prior for $\Wmat$ \cite{Arngren2010}:
\begin{align}
 \pdf(\Wmat) & \propto\expf{-\hyperWgamma \sum_{k=1}^K \norm{\wv_k - \frac{1}{K} \sumKt \wv_{k'} }_2^2 } \positivefun{\Wmat}\!\!,
\end{align}
where $\wv_k$ is the $k$th row of $\Wmat$ and $\positivefun{\cdot}$ returns one if all elements of the argument are positive and zero otherwise. 

The parameter $\hyperWgamma$ needs to be set \textit{a priori} since it cannot be efficiently inferred from the observations.
This is due to the fact that the normalization of $\pdf(\Wmat)$ cannot be computed analytically, such that the conditional $p(\hyperWgamma\,|\,-)$ cannot be derived, which is required for sampling. 

Despite this drawback, this prior has proven to provide a more accurate model than an exponential prior which, in contrast, would allow for efficient sampling of its hyperparameter.

The feature activation matrix $\Amat$ is modeled as \ac{IBP} as described in \secref{sec::ibp}.

\subsection{Posterior model}
From the graphical model depicted in \figref{fig::model::graphicalmodel}, we can derive the posterior given the observed spectra $\Zmat$.
Thus, the joint posterior distribution can be factorized as
\begin{align}
 \begin{split}
 \pdf(\Wmat, \, & \Amat, \Smat, \sigmaZ^2, \hypertauA,\hypertauB, \hyperAalpha, \hyperAbeta\,|\, \Zmat) \propto
 \\ & \, \pdf(\Zmat\,|\, \Wmat, \Amat, \Smat, \sigmaZ^2) \pdf(\sigmaZ^2 \,|\,\hypertauA, \hypertauB)
  \\ & \times \pdf(\Smat) \pdf(\Wmat) \pms(\Amat \,|\, \hyperAalpha, \hyperAbeta) 
 \\ & \times \pdf(\hypertauA) \pdf(\hypertauB) \pdf(\hyperAalpha) \pdf(\hyperAbeta).
 \end{split} \label{eq::model::posterior}
\end{align}

\section{Inference}
\label{sec::inference}
Since an analytical solution of the joint posterior in \eeqref{eq::model::posterior} is not tractable, we represent the posterior by samples generated by means of Gibbs sampling \cite{Geman1984}. 
Therefore, we need to find expressions for the conditionals of the variables.
For convenience, we use the bar symbol ($-$) to denote the set of conditional variables, \ie all variables except the one that is sampled.

\subsection{Sampling the noise variance $\sigmaZ^2$}
The hyperpriors for $\hypertauA$ and $\hypertauB$ are conjugate to the prior. 
Hence, the conditional $\pdf(\sigmaZ^2 \,|\, -)$ is also Inverse-Gamma distributed,
\begin{align*}
 \pdf(\sigmaZ^2 \,|\, -)
        & \propto \pdf(\Zmat\,|\, \Wmat, \Amat, \Smat, \sigmaZ) \pdf(\sigmaZ^2\,|\,\hypertauA, \hypertauB)             
      \\ & \propto \invgampdf{\sigmaZ^2}{\hypertauA + \frac{N D}{2} ,  \right.
         \\ & \quad \left. \hypertauB + \frac{1}{2} \sumN \sumD \left( z_{n,d} - \sumK s_{n,k} a_{k,d} w_{k,d} \right)^2}\!\!,
\end{align*}
where sampling from an Inverse-Gamma distribution is straightforward. 
For the hyperparameters, $\hypertauA$ and $\hypertauB$, the conditionals are
\begin{align*}
 \pdf(\hypertauA \,|\, -)
        & \propto \pdf(\sigmaZ^2\,|\,\hypertauA, \hypertauB) \pdf(\hypertauA\,|\, \hyphypSigmaAlphaA, \hyphypSigmaAlphaB)
     \\ & \propto \invgampdf{\sigmaZ^2}{\hypertauA, \hypertauB} \gampdf{\hypertauA}{\hyphypSigmaAlphaA, \hyphypSigmaAlphaB}\!\!,
\end{align*}
and, analogously,
\begin{align*}
 \pdf(\hypertauB \,|\, -)
        & \propto \pdf(\sigmaZ^2\,|\,\hypertauA, \hypertauB) \pdf(\hypertauB\,|\, \hyphypSigmaBetaA, \hyphypSigmaBetaB)
     \\ & \propto \invgampdf{\sigmaZ^2}{\hypertauA, \hypertauB} \gampdf{\hypertauB}{\hyphypSigmaBetaA, \hyphypSigmaBetaB}\!\!.
\end{align*}
We use an independent Metropolis-Hastings algorithm with a Gaussian proposal distribution to generate samples of $\hypertauA$ and $\hypertauB$.

\subsection{Sampling the abundances $\Smat$}
Since the prior for the abundances imposes a (constrained) uniform distribution of the abundances, the conditional $\pdf(\Smat\,|\,-)$ is proportional to the likelihood in \eeqref{eq::model::likelihood} if the additivity and positivity constraints are fulfilled, and zero otherwise.
Assuming that these constraints hold, we can write the conditional as a Gaussian distribution \cite{Arngren2010}:
\begin{align*}
   \pdf(\sv_{n}\,|\,-) & \propto \expf{-\frac{1}{2\sigmaZ^2} \sumD \left( z_{n,d} - \trans{\vs f_d} \vs s_{n}  \right)^2 }
                  \\ & \propto \gausspdf{\sv_n}{\vs \mu_{\sv_n}}{\vs \Sigma_{\sv_n}}\!\!,
\end{align*}
with mean $\vs \mu_{\sv_n}$ and covariance matrix $\vs \Sigma_{\sv_n}$, 
\begin{align*}
 \vs \mu_{\sv_n}      & = \inv{\left(\sumD \vs f_d \trans{\vs f_d}\right)} \sumD \trans{\vs f_{d}} z_{n,d},\\
 \vs \Sigma_{\sv_n} & = \sigmaZ^2 \Imat_K \inv{\left(\sumD  \vs f_d \trans{\vs f_d}\right)}\!\!\!\!\!\!\!,
\end{align*}
with $\Imat_K$ denoting the identity matrix of size $K$. Thus, we need to sample from a multivariate Gaussian under the constraints that $\sumK s_{n,k} = 1 $ and $0 \leq s_{n,k}$ for all $k = 1,\ldots,K$. 
Sampling from a constrained multivariate Gaussian can be accomplished by Gibbs Sampling \cite{Dobigeon2009}.
Note that we require the hyperparameters of $p(\Smat)$, $\alpha_{s,k}$ with $k=1,\ldots,K$, to be one, otherwise the prior is no longer a (constrained) uniform distribution and sampling the conditional $\pdf(\sv_{n}\,|\,-)$ needs to be conducted by less efficient Metropolis-Hastings sampling.  

\subsection{Sampling the endmember weights $\Wmat$}

The conditional of $\Wmat$, $\pdf(\Wmat\,|\,-)$, is proportional to the likelihood and the prior, \ie
\begin{align*}
 \pdf(\Wmat\,|\,-) & \propto \expf{ -\frac{1}{2\sigmaZ^2} \sumN \norm{ \vs z_{n} - \sumKt s_{n,k'} (\vs a_{k'} \odot \vs w_{k'}) }_2^2
       \\ & \quad -\hyperWgamma \sumKt \norm{ \vs w_{k'} - \frac{1}{K} \sumKtt \vs w_{k''} }_2^2 } \, \positivefun{\Wmat}\!\!.          
\end{align*}
In \cite{Arngren2010}, it is shown that the conditional of the $k$th feature weight vector $\wv_k$ is thus given as
\begin{align*}
 \pdf(\wv_k\,|\,-) & \propto \truncgausspdf{\wv_k}{\vs \mu_{\wv_k}, \vs \Sigma_{\wv_k}}\!\!.
\end{align*}
Recalling that $\fv_k = \av_k \odot \wv_k$ is an element-wise multiplication, the conditional covariance matrix, $\vs \Sigma_{\wv_k}$, and the mean vector, $\vs \mu_{\wv_k}$, can be expressed as
\begin{align*}
  \vs \Sigma_{\wv_k}^{-1} & =  \frac{1}{\sigmaZ^2} \sumN s_{n,k}^2  \diag{\vs a_k} + 2 \hyperWgamma \left(1 - \frac{1}{K} \right) \Imat_D,
  \\ \vs \mu_{\wv_k} &= \vs \Sigma_{\wv_k}^{-1} \Bigg( \frac{1}{\sigmaZ^2} \sumN s_{n,k} ( \vs z_{n} - \sumKtwok s_{n,k'} ( \vs a_{k'} \odot \vs w_{k'}) )^\text{T} \odot \vs a_{k} 
      \\  & \quad - \hyperWgamma \frac{2}{K} \sumKtwok \vs w_{k'} \Bigg).
\end{align*}
Due to the positivity constraint, $\pdf(\wv_k\,|\,-)$ takes the form of a truncated Gaussian. For sampling from a truncated Gaussian, we use the method described in \cite{Chopin2010}.

\begin{algorithm}
  \caption{Sampling new endmembers using an \ac{IBP} prior}
  \vspace{-.5em}
  \label{alg::inference::ibp}
  \begin{tabbing}  
  \hspace{0cm} \= \hspace{1em} \= \hspace{1em} \= \hspace{1em} \= \hspace{1em} \kill
  \\ \> \textbf{for} $d \in 1,\ldots,D$:
  \\ \> \> \textbf{for} $k \in 1,\ldots,K$:
  \\ \> \> \> $a_{k,d} \sim \pms(a_{k,d}\,|\,-)$
  \\ \>  \> $K^+ \sim \pms(K^+\,|\, -)$ 
  \\ \> \> \textbf{for} $k \in 1,\ldots,K^+$:
  \\ \> \> \> $a_{k,d}^+ \leftarrow 1$
  \\ \> \> \> $\wv_{k}^+ \sim \pdf(\Wmat)$
  \\ \> \> \> \textbf{for} $n \in 1,\ldots,N$:
  \\ \> \> \> \> $s_{n,k}^+ \sim \gampdf{}{\frac{1}{K},1}$    
  \\ \> \> $r_\text{accept} \leftarrow $ (\cf \eeqref{eq::inference::ibp::ratio_aug})
  \\ \> \> $P \sim \uniformpdf{}{0,1}$
  \\ \> \> \textbf{if min}($1, r_\text{accept}) < P$:
  \\ \> \> \> $\Smat \leftarrow \text{normalize}([\Smat \ \Smat^+])$
  \\ \> \> \> $\Wmat \leftarrow [\Wmat; \ \Wmat^+]$, $\Amat \leftarrow [\Amat; \ \Amat^+]$  
  \end{tabbing}
%   \end{center}
\end{algorithm}

\subsection{Sampling the endmember activations $\Amat$}
Sampling with an \ac{IBP} prior consists of two steps. 
First, the active columns are updated, \ie the $d$th band of the $k$th endmember is set active with probability 
\begin{align}
  \pms(a_{k,d} = 1\,|\,-) \propto \pdf(\zv_d \,|\, \Smat \fv_d, \sigmaZ^2) \pms(a_{k,d} = 1\,|\, \vs a_{k\backslash d} ).
\end{align}
Second, new features are proposed using a Metropolis step \cite{Doshi-Velez2009, Knowles2011}. 
Assuming fixed means for the prior of $\Wmat$, the proposal distribution, $q(\theta^+ \,|\, \theta)$, for activating $K^+$ endmembers for the $d$th band, is composed of the priors of the latent endmembers and abundances. Hence, the proposal distribution is given as
\begin{align}
  q(\theta^+ \,|\, \theta) = q(\theta^+) = \pms(K^+\,|\, -) \pdf(\Wmat) \pdf(\Smat),
\end{align}
with $\theta = \{ \Wmat, \Amat, \Smat \}$ and $\theta^+ = \{ \Wmat^+, \Amat^+, \Smat^+ \}$ where $\Wmat^+, \Amat^+$, and $\Smat^+$ describe the proposed additional endmember weights, activations, and their abundances.
The acceptance ratio $r$ is given by
\begin{align*} 
 r = \frac{ \pdf(\theta^+ \,|\, \Zmat ,-) q(\theta\,|\,\theta^+ )}{ \pdf( \theta \,|\, \Zmat ,- ) q(\theta^+\,|\,\theta) } = \frac{ \pdf(\Zmat \,|\, \theta^+ ,- ) \pdf(\theta^+) q(\theta\,|\,\theta^+ )}{ \pdf( \Zmat \,|\, \theta  ,-)  \pdf(\theta) q(\theta^+\,|\,\theta) }.
\end{align*}
This expression can be simplified, since $q(\theta\,|\,\theta^+ ) = \pdf(\theta)$. 
The acceptance ratio $r$ is then given by the ratio of the likelihoods only \cite{Meeds2006}: 
\begin{align} 
 r = \frac{ \pdf(\Zmat \,|\, \theta^+  ,-)}{ \pdf( \Zmat \,|\, \theta  ,- )}.
 \label{eq::inference::ibp::ratio}
\end{align}
Note that we run a Gibbs sampler to sample from $\pdf(\Wmat)$ since sampling from this distribution directly is not possible due to the unknown normalization. 
The new abundances, $\Smat^+$, are sampled from a Gamma distribution with parameters $\frac{1}{K}$ and 1. 
We choose these parameters such that the mean of  the proposal for $\Smat^+$ is equal to the mean of the already existing elements in $\Smat$. 
After concatenating the new and existing abundances, the rows of the abundance matrix are normalized to sum to one. This is in line with sampling from a Dirichlet distribution \cite{Devroye1986}.

Following the hints in \cite{Knowles2011}, we augment the ratio in \eeqref{eq::inference::ibp::ratio} with probability $P^+$ of accepting a single new feature, yielding the augmented ratio,
\begin{align}
 r_{\text{aug}} = r \cdot  \frac{  \pms(K^+\,|\, -) }{P^+ \indicatorfun{K^+, 1} + (1 - P^+)  \pms(K^+\,|\, -) },
 \label{eq::inference::ibp::ratio_aug}
\end{align}
with the indicator function $\indicatorfun{a, b}$ returning one if $a$ and $b$ are equal and zero otherwise. 
This increases the probability of proposing new endmembers, leading to faster convergence to the stationary distribution of the Markov chain.
The hyperparameters $\hyperAalpha$ and $\hyperAbeta$ are sampled as described in \secref{sec::ibp::sampling::hyper}. 
The algorithm for sampling new endmembers is outlined in \algref{alg::inference::ibp}.

\begin{figure*}  
  \includegraphics{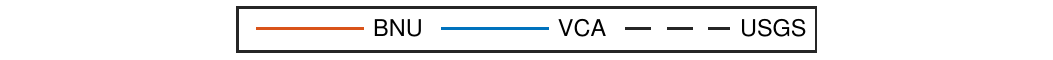} 
  \centering
  \begin{tikzpicture}           

    \node[]                              (m10) {\includegraphics{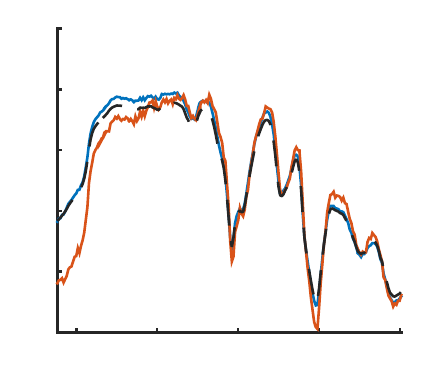}};                                                                               
    \node[right =-.03\textwidth of m10] (m11) {\includegraphics{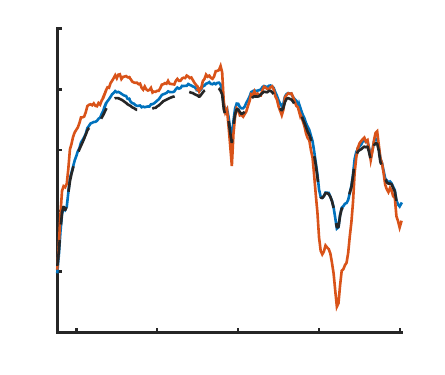}};
    \node[right =-.03\textwidth of m11] (m12) {\includegraphics{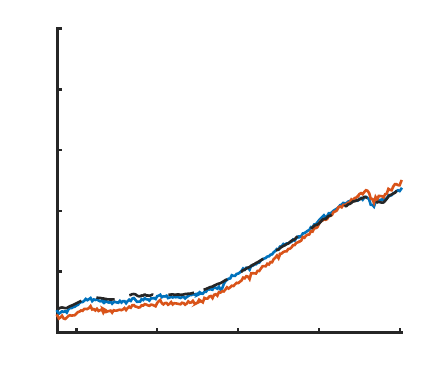}}; 		
    \node[right =-.03\textwidth of m12] (m13) {\includegraphics{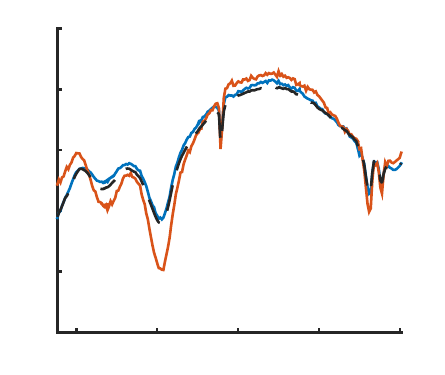}};    
    
    \node[below = 0cm of m10]  (m20) {(a) Carnallite};
    \node[below = 0cm of m11]  (m21) {(b) Ammonioalunite};         	             
    \node[below = 0cm of m12]  (m22) {(c) Biotite};  
    \node[below = 0cm of m13]  (m23) {(d) Actinolite};   
    
  \end{tikzpicture}  
  
  \begin{tikzpicture}             
    \node[] (m30) {\includegraphics{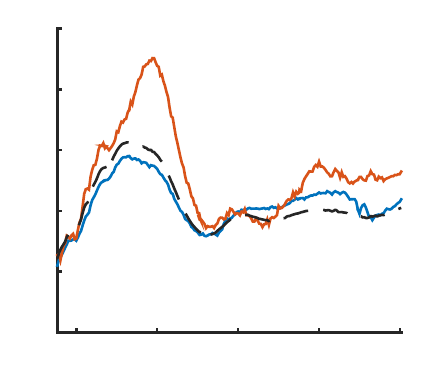}};
    \node[right =-.025\textwidth of m30] (m31) {\includegraphics{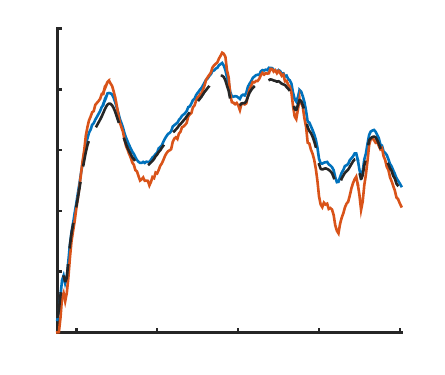}};
    \node[right =-.025\textwidth of m31] (m32) {\includegraphics{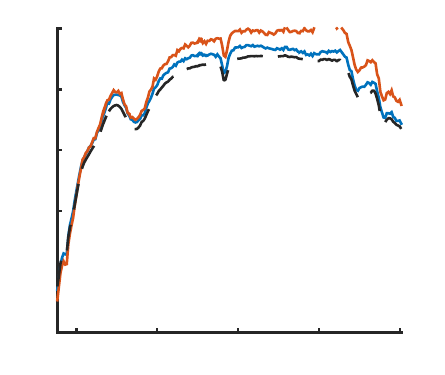}};
    \node[right =-.025\textwidth of m32] (m33) {\includegraphics{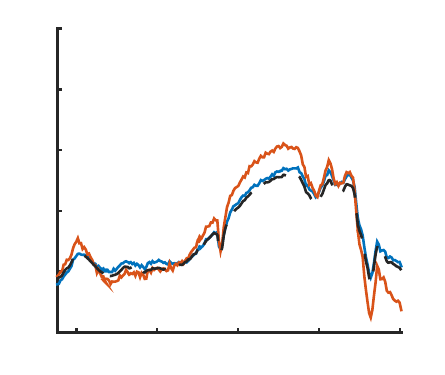}}; 

    \node[below = 0cm of m30]  (m24) {(e) Almandine};
    \node[below = 0cm of m31]  (m41) {(f) Ammonio-jarosite};         	            
    \node[below = 0cm of m32]  (m42) {(g) Andradite};
    \node[below = 0cm of m33]  (m43) {(h) Antigorite};

  \end{tikzpicture} 
  
  \begin{tikzpicture}             
    \node[] (m30) {\includegraphics{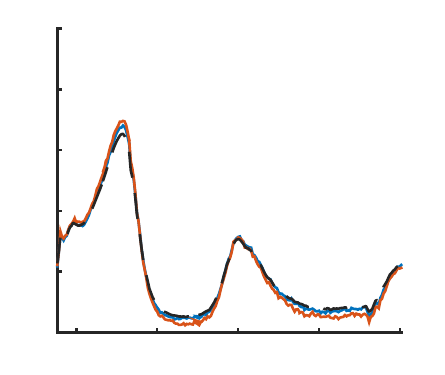}};
    \node[right =-.03\textwidth of m30] (m31) {\includegraphics{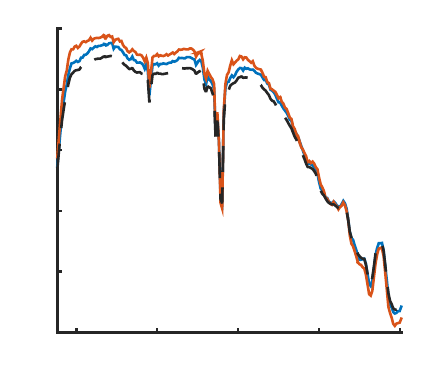}};
    \node[right =-.025\textwidth of m31] (m32) {\includegraphics{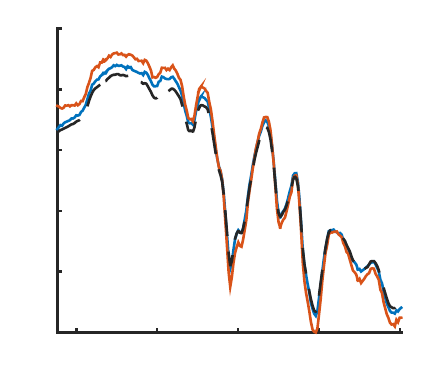}}; 		
    \node[right =-.025\textwidth of m32] (m33) {\includegraphics{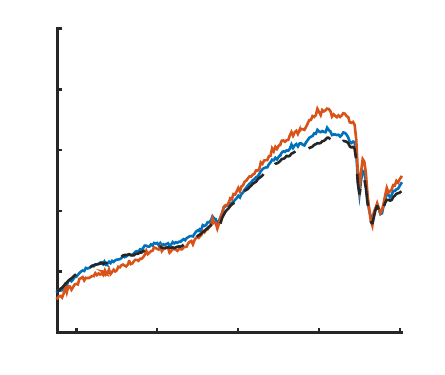}}; 	

    \node[below = 0cm of m30]  (m24) {(i) Axinite};
    \node[below = 0cm of m31]  (m41) {(j) Brucite};         	             
    \node[below = 0cm of m32]  (m42) {(k) Carnallite};
    \node[below = 0cm of m33]  (m43) {(l) Chlorite};    
  \end{tikzpicture}

  \caption{Signatures selected from the USGS spectral library \cite{Clark2007}. For the simulations, the first $K$ signatures are considered as the endmembers of the simulated hyperspectral image. For comparison, examples of endmembers extracted from a simulated hyperspectral image with an \ac{SNR} of \unit{30}{dB} by \ac{VCA} \mbox{($\RMSE{\SID}=0.00243, \RMSE{\text{F}}=1.78, \RMSE{\text{S}}=23.5349$)} and \ac{BNU} ($\RMSE{\SID}=0.0203, \RMSE{\text{F}}=5.32, \RMSE{\text{S}}=13.694$) are depicted. The $x$-axis represents the spectral range from \unit{0.38}{\micro\meter} to \unit{2.5}{\micro\meter} and the $y$-axis denotes the normalized reflectance.}
  \label{fig::sim::signatures}
\end{figure*}

\subsection{Sampling procedure}
\label{sec::inf::sampling}
We start sampling with one feature, \ie we initially set $K=1$.
The first sample of the variables is drawn from the prior distributions. 
As common in Gibbs sampling, the first samples are ignored, as several iterations are needed until the Gibbs sampler generates samples from the target distribution. 

In contrast to other \ac{HSU} algorithms, with a certain probability, new endmembers are introduced in every iteration.
Since we cannot enforce dissimilarity between the proposed and existing endmembers, there is the possibility that new endmembers converge to already present ones, increasing the number of endmembers unnecessarily. 
To alleviate this problem, we could make proposals for merging every combination of the sampled endmembers.
This would, however, lead to the problem that newly created endmembers are easily removed as they have been sampled from the prior (irrespective of the likelihood) and, thus, they basically present noise.

\begin{figure*}  
  \centering
  \includegraphics{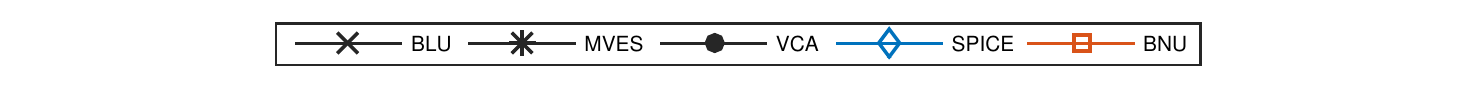} 
  \begin{tikzpicture}
    \node[]                             (m10) {\includegraphics{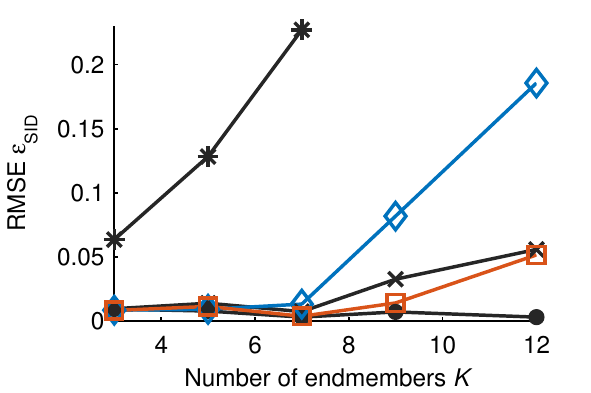}};                                                                               
    \node[right =-.03\textwidth of m10] (m11) {\includegraphics{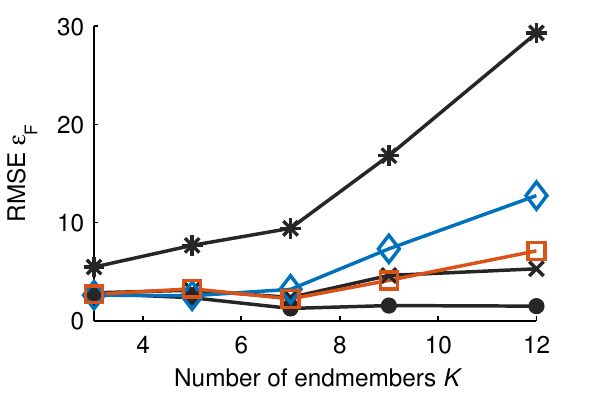}};
    \node[right =-.03\textwidth of m11] (m12) {\includegraphics{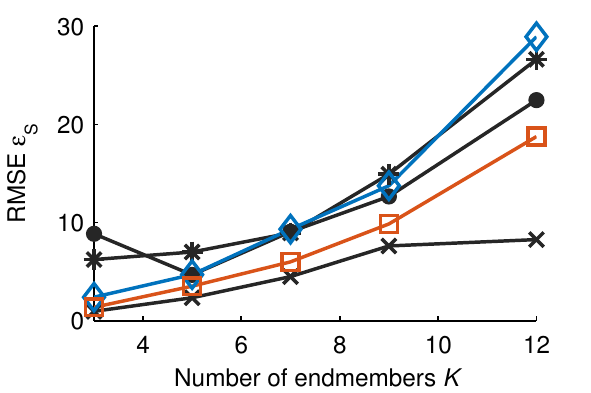}};
  \end{tikzpicture}

  \caption{Simulation results for different numbers of endmembers from 20 Monte Carlo runs, $\SNR = \unit{30}{dB}$. All algorithms perform in a similar range, except \ac{MVES}. \ac{BNU} clearly outperforms \ac{SPICE} and provides even for large number of endmembers good results, close to state-of-the-art algorithms. Results with $\RMSE{\SID} > .25$ are not shown for better comparison.}
  \label{fig::res::unmix::k}
\end{figure*}

For this reason, we consider the following endmember merging strategy: Similar endmembers are likely to have a high probability of being merged. 
Thus, we propose to merge only endmembers that exhibit a correlation above a predefined threshold, $T_\text{corr}$, which also saves computation time. Note that the sampled endmembers are likely to be similar when they are sampled from the stationary distribution. 
Consequently, this scheme prevents merging endmembers too early.
Hence, the described scheme can be understood as an approximation of the combinatorial endmember merging strategy explained above. 
The endmember fusion proposal is then accepted or rejected by means of a Metropolis step, similar as in reversible-jump \ac{MCMC} methods \cite{Green1995}.
Eventually, the activation matrix is likely to become dense, since activations of the merged endmembers are maintained.

It is well known that the Gibbs sampler performs poorly for multi-modal distributions, where the modes are well separated. 
If the density between the modes is close to zero, the Gibbs sampler may not be able to jump between the modes. 
Consequently, the sampler is not able to sample correctly from the distribution.
We observed this behavior in some of our experiments, especially for large number of endmembers.
If we have prior knowledge about the scene, we can adapt the hyperparameters of the model such that the sampler is initialized closer to the target distribution. 
Exploiting prior knowledge, however, still does not guarantee correct sampling of the posterior. 
A better solution is the use of parallel tempering (PT) \cite{Earl2005}. 
Here, the idea is to run multiple Markov chains in parallel at different temperatures. 
A Metropolis step is then introduced for swapping the states of the chains.
In our model, the temperature can be understood as an additional variance term on the likelihood, \ie the likelihood is smoothed. 
Thus, chains at higher temperatures (higher variances) are likely to overcome modes of the posterior. 
Note that the first chain always samples at temperature $T_{\text{PT}}= 1$, generating valid samples of the target distribution. 
Making swap proposals only every few iterations allows to run the algorithm on a parallel architecture. 
We cool down the temperatures of all chains, similar as in simulated annealing \cite{Kirkpatrick1983,vCerny1985}, to ensure that all chains are swapped after several iterations.

After several iterations of the sampler, an approximation of the \ac{MAP} estimate of the endmembers, $\hat{\Fmat}$, and the abundances, $\hat{\Smat}$, is given by 
the sample with the highest posterior probability. 
The posterior probability can be calculated from \eeqref{eq::model::posterior}. Note that the obtained estimate is effectively simply a realization of the random variables.
This will be discussed in \secref{sec::discussion}.

\section{Experimental Results}
\label{sec::results}
We compare our proposed algorithm, \ac{BNU}, with different state-of-the-art unmixing algorithms.
For geometrical based algorithms, we consider \acf{VCA} \cite{Nascimento2005}, \acf{MVES} \cite{Chan2009}, and \acf{SPICE} \cite{Zare2007}. 
Further, we investigate the Bayesian approach presented in \cite{Dobigeon2009}, \acf{BLU}.

As none of these methods, except \ac{SPICE}, is able to estimate the number of endmembers, we also provide results for the following endmember dimensionality estimation algorithms: \acf{VD} \cite{Chang2004}, \acf{HySime} \cite{Bioucas-Dias2008}, \acf{hideNN} \cite{Heylen2013}, and \ac{SPICE}. 

\begin{table}
 \caption{Parameters used for \ac{BNU} in the simulation experiments}
 \label{tab::res::params}
 \begin{tabular}{r|c|l}
     Parameter & Value & Meaning \\  \hline \hline
   $\hyphypSigmaAlphaA, \hyphypSigmaAlphaB,$  & \multirow{ 2}{*}{\!\!\!\!\!\Big\} 1} &  \multirow{ 2}{*}{hyperparameters for $\sigmaZ^2$}
  \\ $\hyphypSigmaBetaA, \hyphypSigmaBetaB$ & 
  \\ $\hyphypAalphaA, \hyphypAalphaB$  & 1 & hyperparameters for $\hyperAalpha$
  \\ $\hyphypAbetaA$  & 1 & hyperparameter for $\hyperAbeta$
  \\ $\hyphypAbetaB$  & 10 & hyperparameter for $\hyperAbeta$
  \\ $\hyperWgamma$     & 100 & weighting of the prior for $\Wmat$
  \\ $P^+$          & 0.1  & probability of accepting $K^+=1$ features
  \\ $T_\text{corr}$ & 0.95 & threshold for merging similar endmembers
  \\ $N_\text{iter}$ & $10,000$ & number of iterations of the Gibbs Sampler
 \end{tabular}
\end{table}

\begin{figure*}  
  \centering
  \includegraphics{su_legend}
  \begin{tikzpicture}
    \node[]                             (m10) {\includegraphics{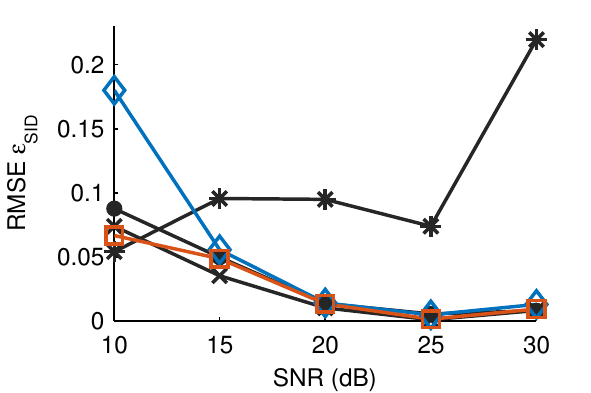}};                                                                               
    \node[right =-.03\textwidth of m10] (m11) {\includegraphics{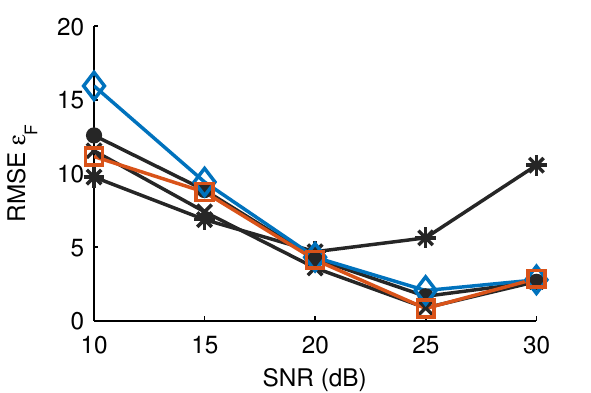}};
    \node[right =-.03\textwidth of m11] (m12) {\includegraphics{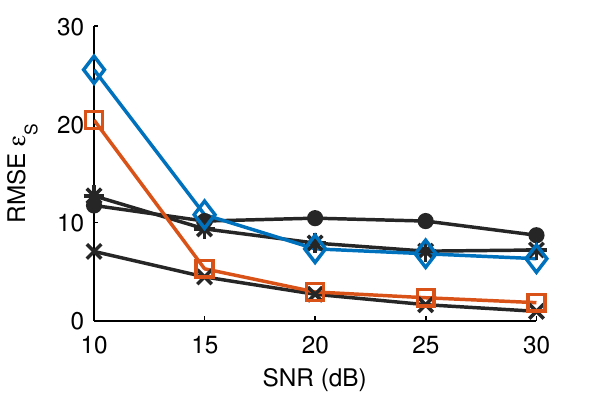}};    
  \end{tikzpicture}

  \caption{Simulation results for different \ac{SNR}s from 20 Monte Carlo runs, $K=3$. \ac{MVES} shows a poor overall performance for endmember extraction, as in some simulation runs it resulted in extremely poor estimates. The other algorithms perform in a similar range, yielding better results with increasing \ac{SNR}. In all scenarios, \ac{BNU} yields better results than \ac{SPICE}, especially in case of low SNR.} 
  \label{fig::res::snr::rmse}
\end{figure*}

In the simulation experiments, or if ground truth data is available, we set the correct number of endmembers for \ac{VCA}, \ac{MVES} and \ac{BLU}. 
In contrast, \ac{SPICE} and \ac{BNU} do not require information about the number of endmembers.
As explained, they aim not only at estimating the endmembers and their fractional abundances but also the number of the endmembers. 
For this reason, we compare \ac{BNU} especially with \ac{SPICE} and highlight performance differences between both methods.

For \ac{BNU}, we choose the parameters shown in \tabref{tab::res::params} unless otherwise stated. 
For the other algorithms, the parameters are set to their default values. Only in case of \ac{hideNN}, we tuned the parameters as the default values lead to poor results in our simulations. 
We set the false alarm rate to $10^{-3}$ for \ac{VD}. 
Due to the underlying assumptions of \ac{hideNN}, this algorithm provides floating point estimates of $K$ which are rounded for comparison.

As figure of merits, we consider three different measures: the average angular difference between the true and estimated endmembers, $\check{\fv}_k$ and $\hat{\fv}_k$ with $k=1,\ldots,K$, 
\begin{align*}
 \overline{\theta}_\text{F} = \frac{1}{K} \sumK \arccosfun{\frac{\check{\fv}_k  \trans{ \hat{\fv}_k} }{ {\norm{ \check{\fv}_k}}_2 {\norm{\hat{\fv}_k}}_2 }}\!\!,
\end{align*}
the average angular difference between the true and estimated abundances, $\check{\sv}_k$ and $\hat{\sv}_k$ with $k=1,\ldots,\numFeats$, 
\begin{align*}
 \overline{\theta}_\text{S} = \frac{1}{\numFeats} \sumK \arccosfun{\frac{\trans{  \check{\sv}_k }\hat{\sv}_k }{ {\norm{\check{\sv}_k}}_2 {\norm{\hat{\sv}_k}}_2 }}\!\!,
\end{align*}
and the average spectral information divergence, $\overline{\text{SID}}$ \cite{Chang2000}. 
The SID measures the difference between the endmembers based on the symmetric Kullback-Leibler divergence \cite{Kullback1968},
\begin{align*}
 \SID_k = \sumD p_{d,k} \logfun{ \frac{p_{d,k}}{q_{d,k}} } + \sumD q_{d,k} \logfun{ \frac{q_{d,k}}{p_{d,k}} }\!\!,
\end{align*}
with $p_{d,k} = \check{f}_{k,d} / \sumDt \check{f}_{k,d'}$ and $q_{d,k} = \hat{f}_{k,d} / \sumDt \hat{f}_{k,d'}$. 
Thus, the average SID is given as $\overline{\SID} = \frac{1}{K}\sumK \SID_k$.
We compute the \ac{RMSE} of each of the measures, $\RMSE{\text{F}}$, $\RMSE{\text{S}}$, and $\RMSE{\SID}$, as in \cite{Nascimento2005}.

In order to evaluate the endmember dimensionality estimation algorithms, we consider the rate of correctly estimated number of endmembers. 
We refer to the average rate of correctly estimated (in the sense of the ground truth) endmembers as \emph{Accuracy}.
We also consider the \ac{RMSE} of the dimensionality estimate $K$, $\RMSE{\text{K}}$, over all Monte Carlo runs, taking into account the amount of error introduced by incorrect estimates. Please note that, in real data, the number of endmembers, $K$, depends on the assumed type of endmembers and their tolerated fractional abundances. 
Therefore, it is often difficult to determine a correct $K$ for real data sets. 

\subsection{Simulations}
In this section, we evaluate the performance of the proposed algorithm with respect to 1) the (latent) number of endmembers, 2) different \ac{SNR}s, and 3) illumination perturbation. We investigate simulations with the following setups.

First, $K \in \{3, 5, 7, 9, 12\}$ pure materials are chosen from the USGS spectral library \cite{Clark2007}.
The chosen endmembers are depicted in \figref{fig::sim::signatures}, along with estimates obtained by \ac{BNU} and \ac{VCA}.
Samples from a Dirichlet distribution, with hyperparameters identically set to $\frac{1}{K}$, are drawn to create the ground truth for the abundances, resulting in a hyperspectral image of $40 \times 40$ pixels and 224 bands. 

Second, we fix the number of endmembers to $K=3$ and apply additive Gaussian noise to the image with different \ac{SNR}s, ranging from \unit{10}{dB} to \unit{30}{dB} with a step size of \unit{5}{dB}. 

Third, the effect of illumination perturbation, \ie varying lighting conditions, is investigated by applying multiplicative Beta distributed noise ($\sim \betapdf{}{\paramIllum,1}$) to the image with parameter $\paramIllum \in \{1, 5, 10, 15, 20, 25, 30\}$.
In these simulations, the number of endmembers is set to $K=3$ and an \ac{SNR} of \unit{30}{dB} is considered.

We present the \ac{RMSE}s over 20 Monte Carlo runs for each simulation.
Especially in the case of low \ac{SNR}s, as will be shown, $K$ is estimated only in few examples correctly. 
Thus, we also include the results when $K$ is overestimated (in case of \ac{SPICE} and \ac{BNU}) to increase the number of samples used to calculate the \ac{RMSE}s. 

A significant drawback of \ac{HySime} and \ac{VCA} is their requirement of knowledge about the noise. 
For \ac{HySime}, the provided noise estimator is utilized, while we provide \ac{VCA} with the exact \ac{SNR}.

\subsubsection{Number of endmembers} 
As can be observed in \figref{fig::res::unmix::k}, \ac{BNU}, \ac{BLU} and \ac{VCA} yield the lowest errors in the reconstruction, where \ac{BNU} clearly outperforms \ac{SPICE}. 
Especially in scenarios with many endmembers, \ac{BNU} provides highly accurate endmember extraction and abundance estimation, exceeded only by \ac{BLU}.
The poor overall performance of \ac{MVES} can be explained by the observation that \ac{MVES} results in extremely poor estimates in some simulation configurations, where the other methods still provide good results.
Note that, in practice the results of \ac{VCA} are likely to be worse since \ac{VCA} is provided with perfect knowledge about the \ac{SNR}. 

In \figref{fig::sim::signatures}, realizations of the extracted endmembers obtained by \ac{BNU} and \ac{VCA} are presented. 
\ac{BNU} and \ac{VCA} show results of similar high accuracy, while \ac{BNU}, as opposed to \ac{VCA}, additionally estimated the number of endmembers.

\begin{figure}
  \centering
  \includegraphics[scale=.5]{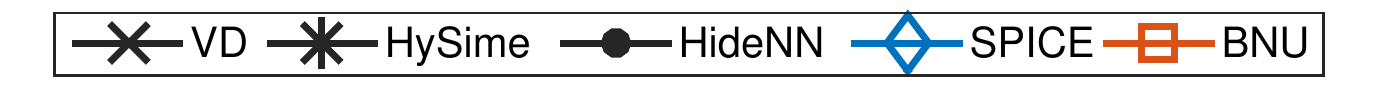}\\
  \begin{tikzpicture}
    \node[]                             (m10) {\includegraphics{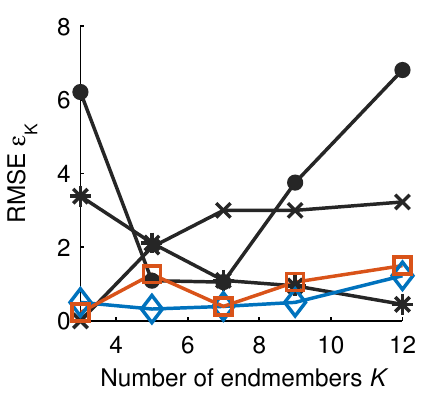}};                                                                               
    \node[right =-.03\textwidth of m10] (m11) {\includegraphics{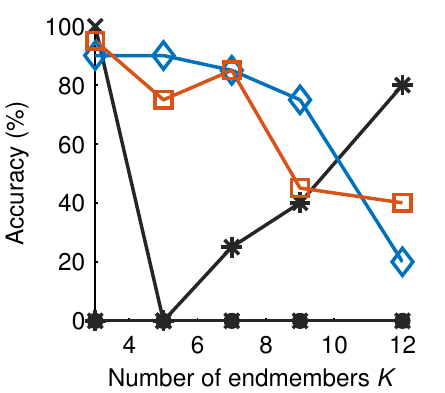}};          
  \end{tikzpicture} 
  \caption{Results of the dimensionality estimation for different number of endmembers, \ac{RMSE}s (left) and accuracies (right), \ac{SNR} = \unit{30}{dB}. \ac{BNU} and \ac{SPICE} provide the most accurate estimates. \ac{VD} and \ac{hideNN} fail at estimating $K$ correctly when the true number of endmembers is greater than 5.}
  \label{fig::res::dim::k}
\end{figure}

\begin{figure*}  
  \centering
  \includegraphics{su_legend} \\
  \begin{tikzpicture}
    \node[]                             (m10) {\includegraphics{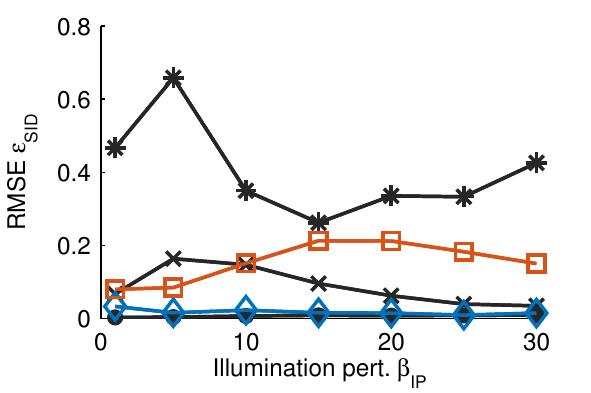}};                                                                               
    \node[right =-.03\textwidth of m10] (m11) {\includegraphics{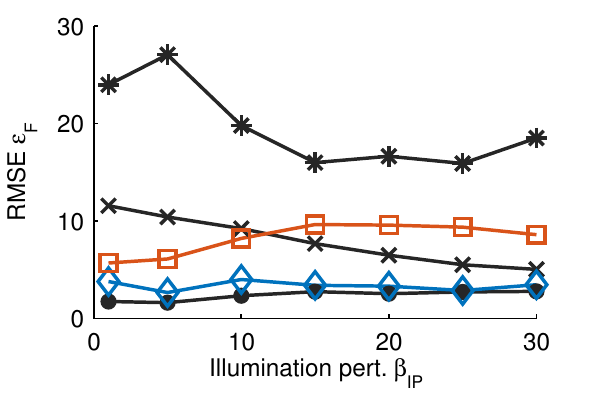}};
    \node[right =-.03\textwidth of m11] (m12) {\includegraphics{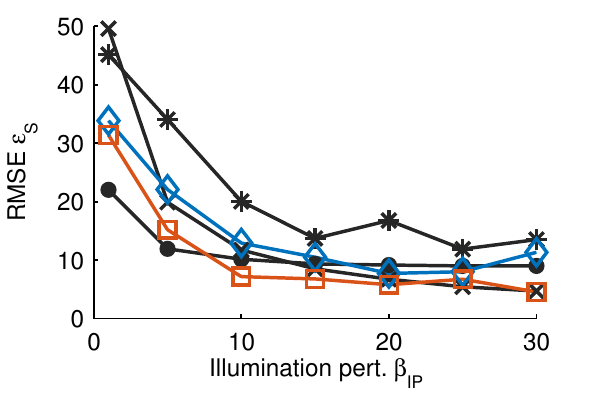}};
  \end{tikzpicture}

  \caption{Simulation results for different illumination perturbations, \ac{SNR} = \unit{30}{dB} and $K$ = 3. Low values of $\paramIllum$ lead to low \ac{SNR}s due to stronger perturbations. All methods except \ac{MVES} yield good results for endmember extraction. The obtained abundance estimates show strong errors in case of low $\paramIllum$ for all methods.  While \ac{SPICE} outperforms \ac{BNU} for endmember estimation, \ac{BNU} shows a better performance for abundance estimation.}
  \label{fig::res::unmix::illum}
\end{figure*}

The results for dimensionality estimation are depicted in \figref{fig::res::dim::k}. \ac{BNU} and \ac{SPICE} yield precise estimates, clearly outperforming the other methods. 
However, their performances decrease towards larger numbers of endmembers. 
\ac{HySime} performs only well for large $K$. \ac{VD}, on the contrary, shows good performance only for few endmembers. 
HideNN results in the highest error for estimating the number of endmembers, despite our attempts to find suitable parameters.
Though showing a low \ac{RMSE} for $K=5$ and $K=7$, \ac{hideNN} fails to correctly estimate $K$ in all simulations. 

\subsubsection{Noise levels}
In order to investigate the effect of noise, different \ac{SNR} levels have been applied to the simulated image. 
As shown in \figref{fig::res::snr::rmse}, concerning endmember extraction, \ac{BNU} provides highly accurate estimates, equal to the other methods. If the observations suffer from strong noise (\ac{SNR} $<$ \unit{15}{dB}), \ac{BNU} clearly outperforms \ac{SPICE}. 

For the estimation of the abundances, \ac{BNU} results, together with \ac{BLU}, in the most accurate estimates. Only for low \ac{SNR}s ($<$ \unit{15}{dB}), BNU is outperformed by \ac{BLU}. \ac{SPICE}, however, performs in all scenarios worse than \ac{BNU}.
 
\begin{figure}
  \centering
  \includegraphics[scale=.5]{dim_legend} \\
  \begin{tikzpicture}
    \node[]                             (m10) {\includegraphics{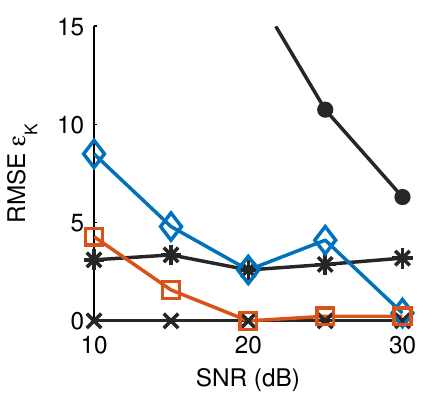}};
    \node[right =-.03\textwidth of m10] (m11) {\includegraphics{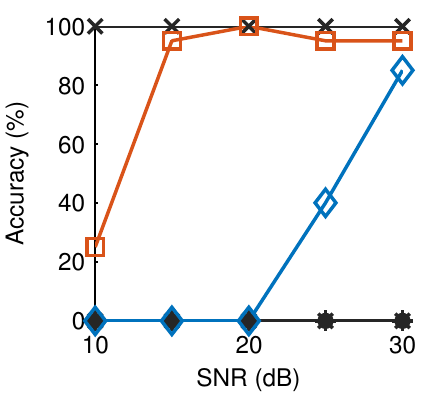}};   
  \end{tikzpicture}  
  \caption{Results of the dimensionality estimation for different \ac{SNR}s, \ac{RMSE}s (left) and accuracies (right), $K=3$. \ac{BNU} and \ac{VCA} provide the best results, significantly outperforming the other methods. Note that \ac{VCA} performs especially well for a low number of endmembers and is likely to show worse results with a large number of endmembers. Results with $\RMSE{\text{K}} > 15$ are not shown for better comparison.}
  \label{fig::res::snr::dim}
\end{figure}

The results of the dimensionality estimation presented in \figref{fig::res::snr::dim} are similar to the results of the previous simulations, showing that \ac{BNU} and \ac{VD} provide high accuracies.
In case of strong noise (\ac{SNR} $<$ \unit{20}{dB}), the accuracy of \ac{BNU} decreases significantly, but is still remarkably higher than of \ac{SPICE}. 
Observing the \ac{RMSE}s, it becomes clear that \ac{BNU} outperforms \ac{SPICE} for all investigated \ac{SNR}s. Only \ac{VD} and \ac{HySime} are able to provide results similar to \ac{BNU} in terms of the \ac{RMSE}.

\subsubsection{Illumination perturbation}
\label{res::sim::illum}
As proposed in \cite{Nascimento2005}, multiplicative noise following a Beta distribution with parameter $\paramIllum$
is applied to the abundances in order to simulate illumination perturbations caused by different lighting conditions. 
Note that the higher $\paramIllum$, the less perturbation is implied, as the probability of sampling a value close to one is increased, resulting in little perturbation only. In contrast, with $\paramIllum=1$, we scale the abundances by a value drawn from a uniform distribution in the range from zero to one, which leads to highly noisy simulated observations. 

From \figref{fig::res::unmix::illum}, we observe that the endmembers are still well extracted by all tested algorithms except \ac{MVES}. \ac{BNU} and \ac{BLU} yield similar results, slightly worse than the performance of \ac{VCA} and \ac{SPICE}.
For the abundance estimation, \ac{BNU} provides the most accurate estimates when $\paramIllum > 5$ and is outperformed only by \ac{BLU} in case of stronger perturbations.
The performance decrease of the Bayesian approaches compared with the results of the previous experiments can be explained by the fact that both generative models do not consider multiplicative noise. 

\begin{figure}
  \centering
  \includegraphics[scale=.5]{dim_legend} \\
  \begin{tikzpicture}
    \node[]                             (m10) {\includegraphics{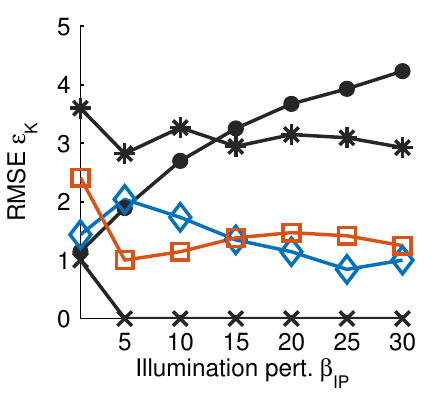}};
    \node[right =-.03\textwidth of m10] (m11) {\includegraphics{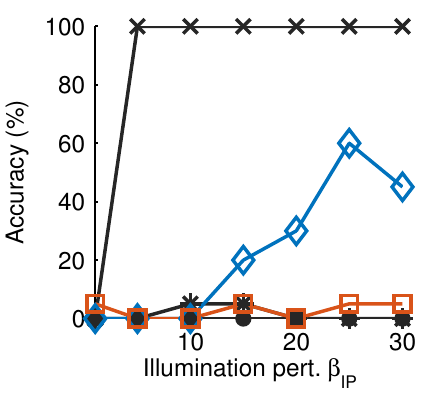}}; 
  \end{tikzpicture}  
  \caption{Results of the dimensionality estimation for different illumination perturbations, \ac{RMSE}s (left) and accuracies (right). Only \ac{VD} is able to estimate the number of endmembers precisely. \ac{BNU} provides still good results, overestimating the number of endmembers often by one, yielding similar results as \ac{SPICE}. The other methods show significant errors.}
  \label{fig::res::illum::dim}
\end{figure}

\begin{figure}
  \centering
  \includegraphics[scale=0.5]{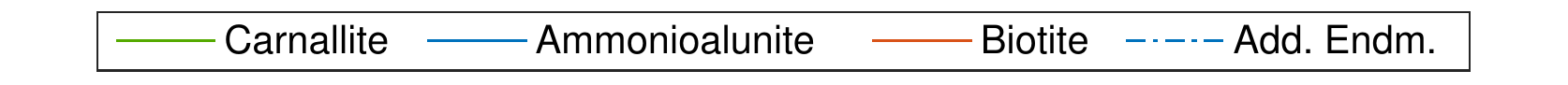} \\
  \includegraphics{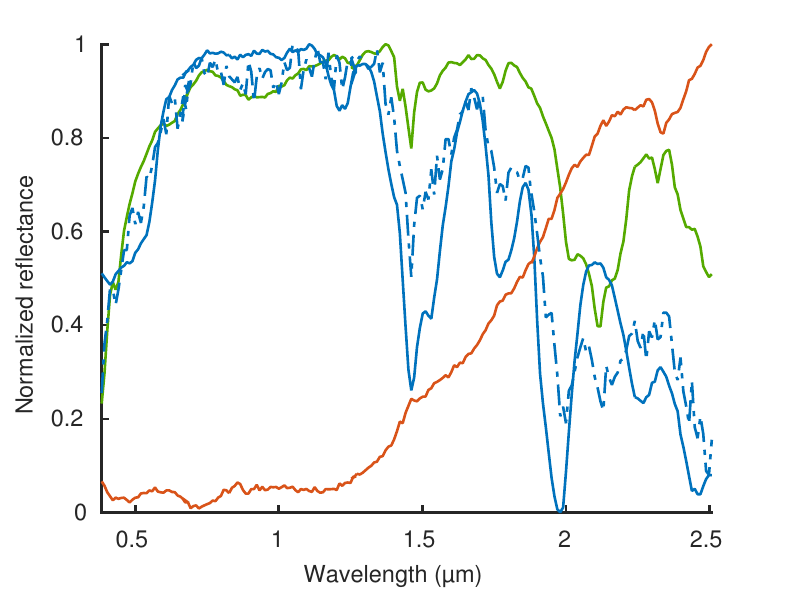}
  \caption{Realization of the endmembers for the illumination perturbation simulation, $\paramIllum = 25$. The additional endmember (dashed line) is basically a noisy replica of Ammonioalunite, absorbing the multiplicative noise of the observations.}
  \label{fig::res::illum::addendmember}
\end{figure}

\figref{fig::res::illum::dim} reveals that none of the examined methods, except \ac{VD}, is able to estimate the numbers of endmembers in this setup reliably.
\ac{SPICE} is able to recover the true value in some cases, as long as the noise is not too strong.
While \ac{BNU}, \ac{HySime}, and \ac{hideNN} fail at estimating the correct dimensionality in most cases, \ac{BNU} overestimates the number of endmembers by only 1 or 2, which explains the low \ac{RMSE}.
The additional endmembers often present (scaled) noisy versions of the endmembers appearing in the scene as illustrated in \figref{fig::res::illum::addendmember}, while the truly present endmembers are well reconstructed.
Thus, the additional endmembers can be understood as noise absorption endmembers, \ie information which cannot be explained by the model is absorbed into additional endmembers in order to explain the observed data. 

\begin{figure}
  \centering
  \includegraphics[width=.35\textwidth]{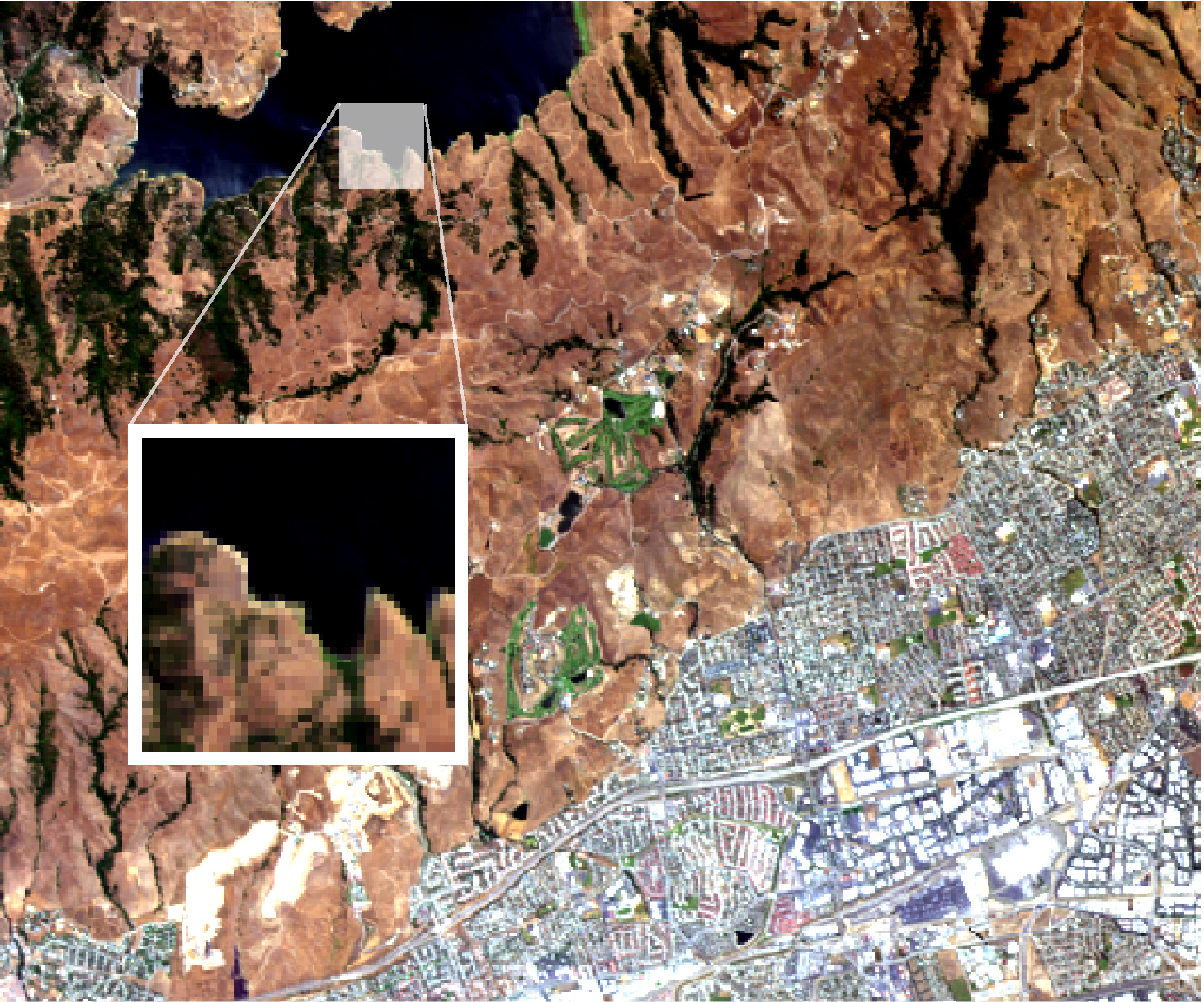}
  \caption{True color image of the Moffett field. The selected subset has a size of $50 \times 50$ pixels and contains different signatures such as soil, vegetation, and water.}
  \label{fig::res::moffett::real}
\end{figure}

\begin{figure}
  \centering
  \includegraphics[width=.35\textwidth]{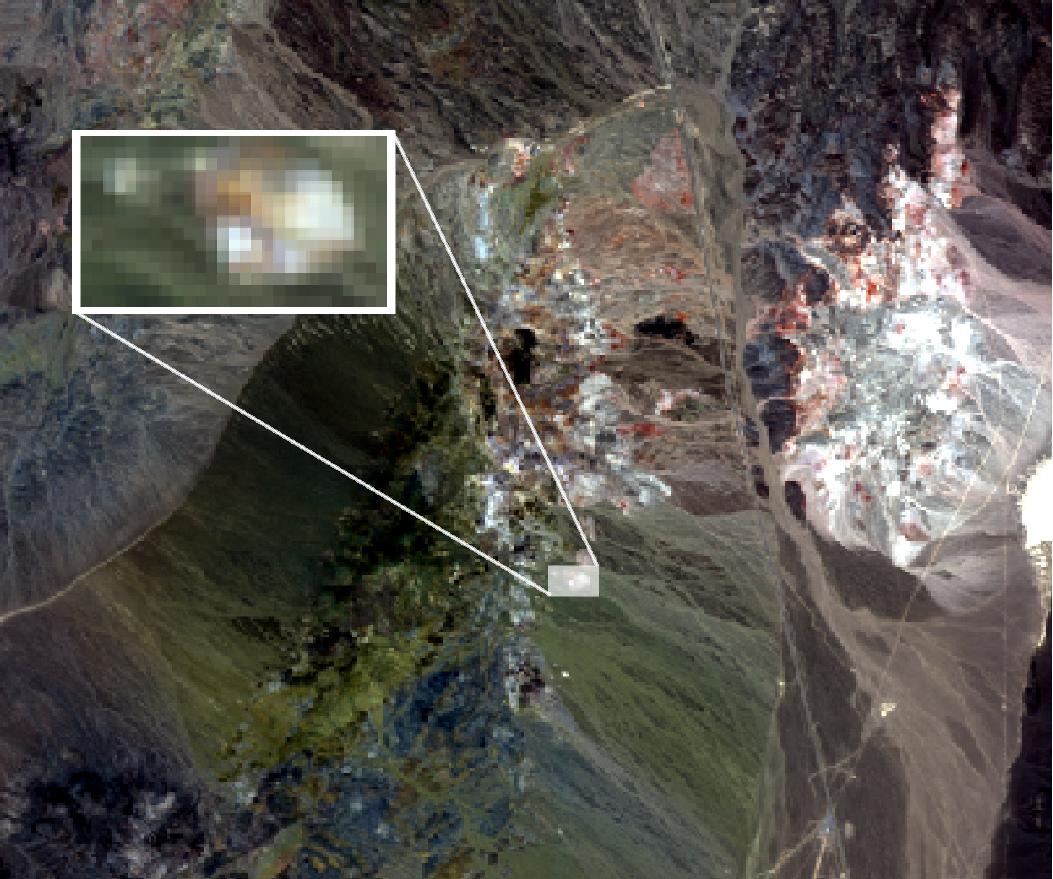}
  \caption{True color image of the Cuprite scene. The selected subset has a size of $16 \times 28$ pixels and contains signatures from various minerals.}
  \label{fig::res::cuprite::real}
\end{figure}

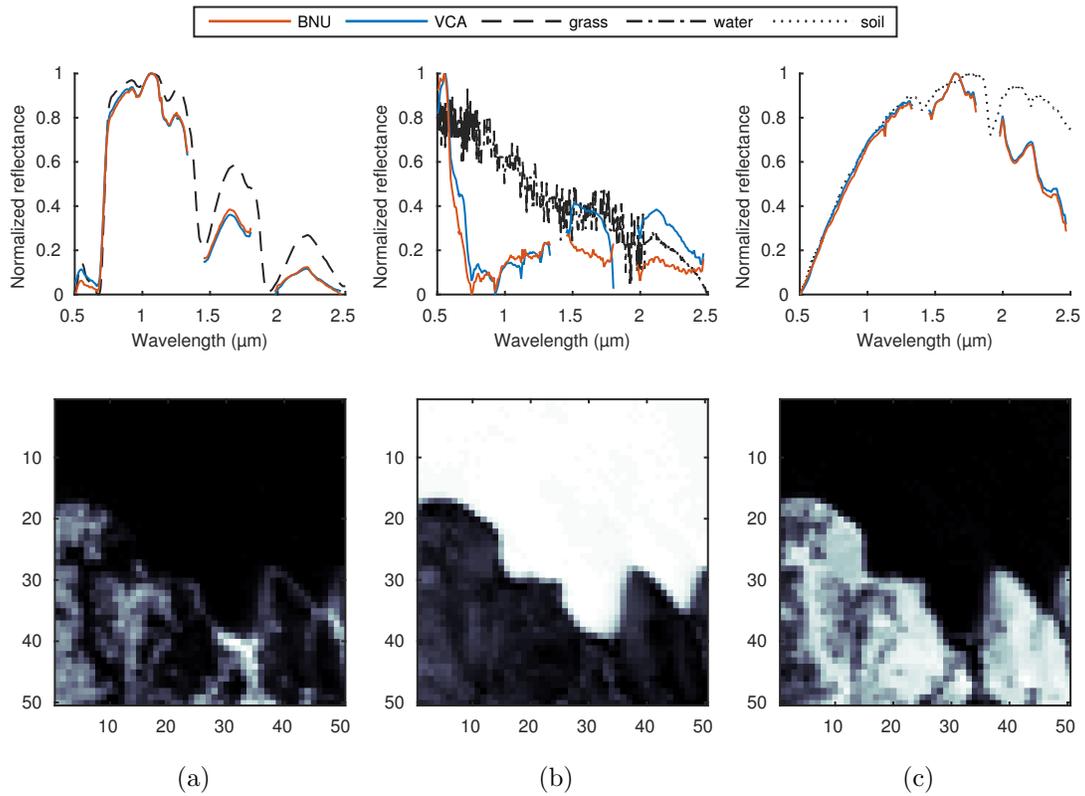
\begin{figure*}  
  \centering
  \includegraphics{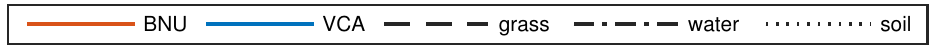} \\
  \begin{tikzpicture}    
    \node[]                             (m10) {\includegraphics{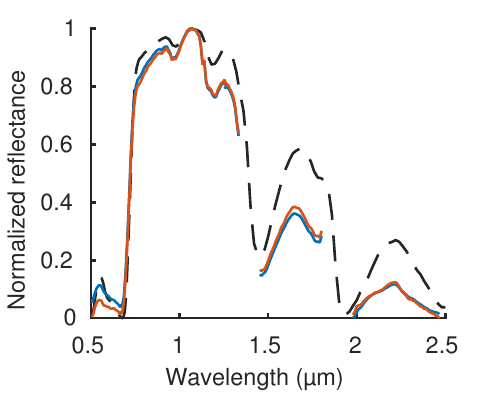}};                                                                               
    \node[right =-.03\textwidth of m10] (m11) {\includegraphics{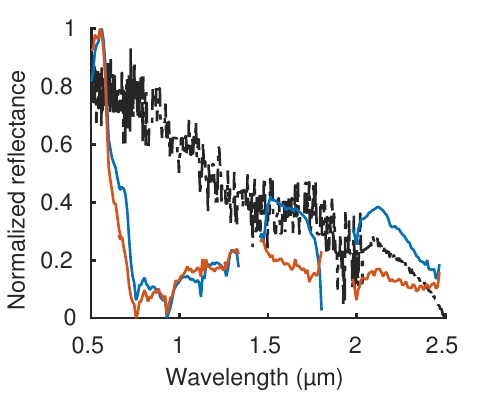}};
    \node[right =-.03\textwidth of m11] (m12) {\includegraphics{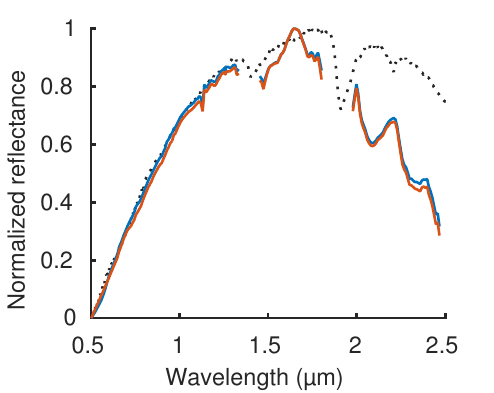}};
         
  \end{tikzpicture}
  \begin{tikzpicture}
    \node[]                             (m10) {\includegraphics{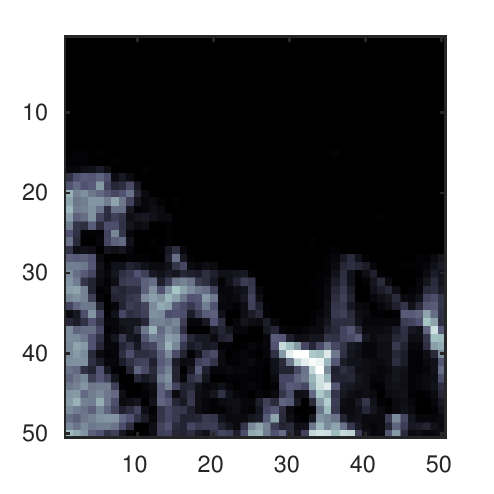}};                                                                               
    \node[right =-.03\textwidth of m10] (m11) {\includegraphics{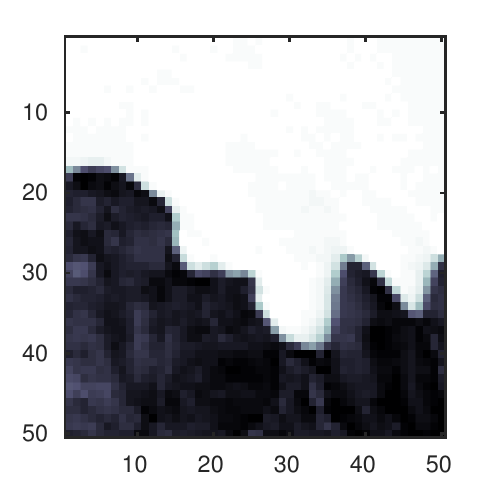}};
    \node[right =-.03\textwidth of m11] (m12) {\includegraphics{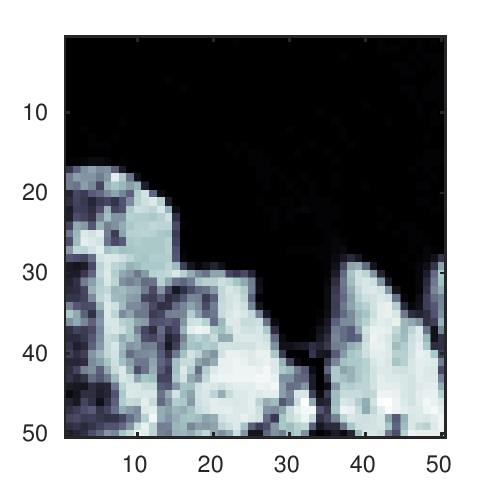}};  
    \node[below = 0cm of m10]  (m20) {(a)};
    \node[below = 0cm of m11]  (m21) {(b)};
    \node[below = 0cm of m12]  (m22) {(c)}; 
  \end{tikzpicture}
  \caption{Results of the Moffett field, extracted endmembers (top) and the estimated abundances (bottom). For the abundances, \emph{white} refers to 1 and \emph{black} to 0. In each of the five Monte Carlo runs, \ac{BNU} extracted three endmembers: (a) grass (vegetation), (b) water, and (c) soil. The estimated abundances represent the scene well and are similar to the results shown in \cite{Dobigeon2009}.} %In the estimated abundance maps, white refers to one, and black to zero.}
  \label{fig::res::moffett::emam}
\end{figure*}

\subsection{Real data}
For real data evaluation, we consider two different subsets of real data sets, 1) the Moffett field and 2) the Cuprite scene. 
While the Moffett field contains rather vegetation and urban signatures, the Cuprite scene shows mainly geological features such as different minerals and rocks.
For comparison, we show the endmembers estimated by \ac{VCA} and the most similar endmembers chosen from the ASTER spectral library \cite{Baldridge2009}. 
Though \ac{BNU} often converges after several hundred iterations, we select the sample maximizing the posterior after $10,000$ samples to ensure convergence of the Markov chain to obtain an approximate \ac{MAP} estimate. 
Five Monte Carlo runs are performed to make sure that the results are independent of the initialization and the randomly generated samples. 
For the Moffett field, we set $\hyperWgamma = 2,000$ and for Cuprite $\hyperWgamma = 500$.
Since no ground truth is available, we use a subset of both data sets as in \cite{Dobigeon2008, Dobigeon2009}, which allows for a more detailed analysis. Further, the \ac{SNR} is assumed to be \unit{30}{dB} for \ac{VCA}.

\begin{figure*}  
  \centering
  \includegraphics{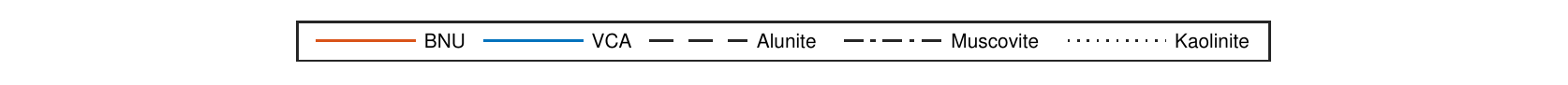} \\
  \begin{tikzpicture}    
    \node[]                             (m10) {\includegraphics{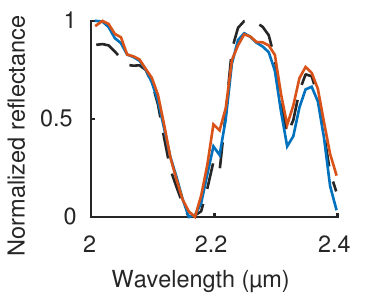}};                                                                               
    \node[right =-.02\textwidth of m10] (m11) {\includegraphics{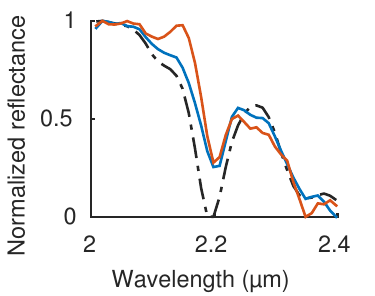}};
    \node[right =-.02\textwidth of m11] (m12) {\includegraphics{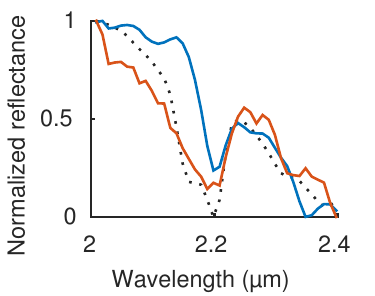}};
    \node[right =-.02\textwidth of m12] (m13) {\includegraphics{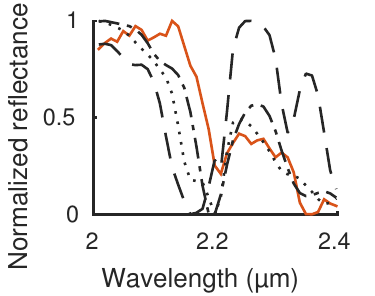}};
    \node[right =-.02\textwidth of m13] (m14) {\includegraphics{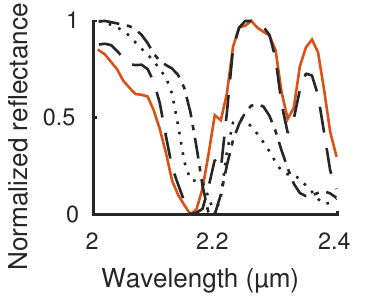}};        
  \end{tikzpicture}
  \begin{tikzpicture}
    \node[]                             (m10) {\includegraphics{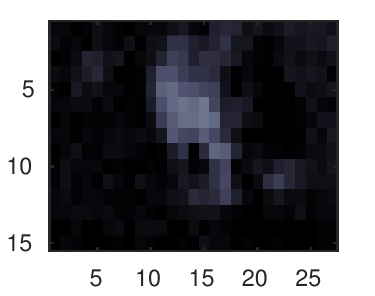}};                                                                               
    \node[right =-.02\textwidth of m10] (m11) {\includegraphics{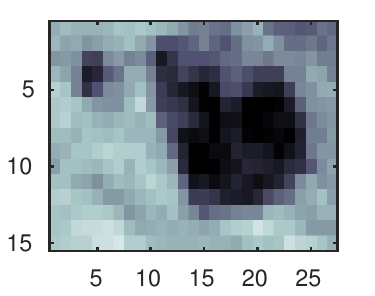}};
    \node[right =-.02\textwidth of m11] (m12) {\includegraphics{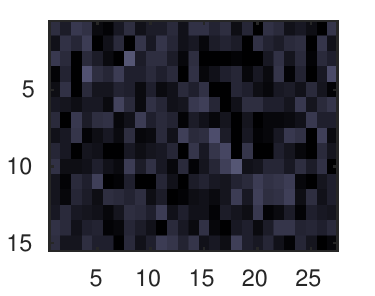}};
    \node[right =-.02\textwidth of m12] (m13) {\includegraphics{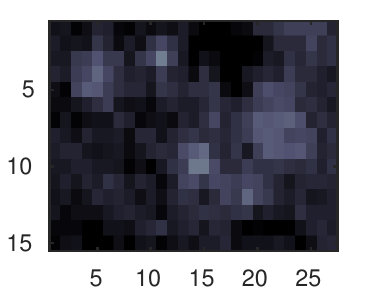}};
    \node[right =-.02\textwidth of m13] (m14) {\includegraphics{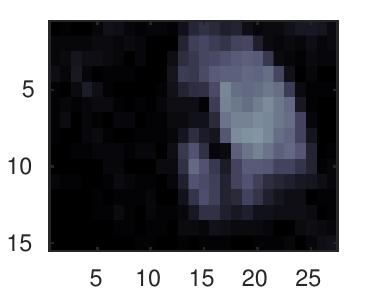}};
    \node[below = 0cm of m10]  (m20) {(a)};
    \node[below = 0cm of m11]  (m21) {(b)};
    \node[below = 0cm of m12]  (m22) {(c)};         
    \node[below = 0cm of m13]  (m23) {(d)};
    \node[below = 0cm of m14]  (m24) {(e)}; 
 
  \end{tikzpicture}
  \caption{Results of the Cuprite scene, extracted endmembers (top) and the estimated abundances (bottom). For the abundances, \emph{white} refers to 1 and \emph{black} to 0. \ac{BNU} samples between four to five endmembers in all five Monte Carlo runs. In the illustrated realization, (a) Alunite, (b) Muscovite, (c) Kaolinite, and noisy copies of (d) Kaolinite and (e) Alunite are extracted. Though in this example, the feature merging proposal is rejected, the results are similar with those reported in \cite{Dobigeon2009}. }
  \label{fig::res::cuprite::emam}
\end{figure*}
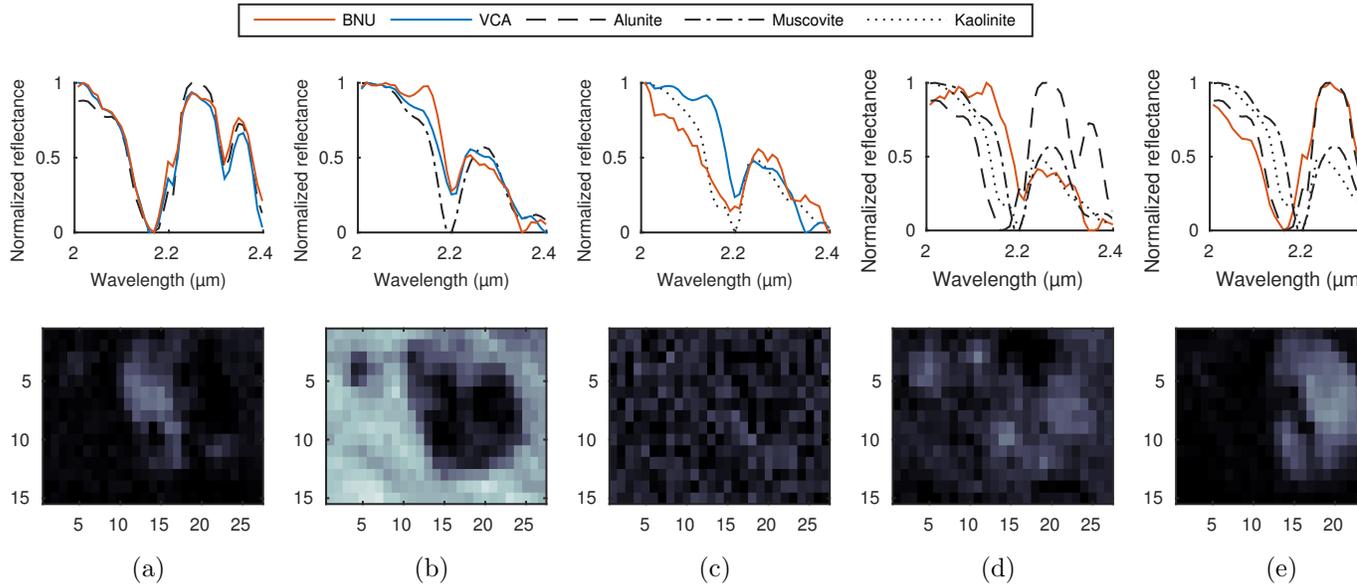

\subsubsection{Moffett field}  
The Moffett field scene was captured over Moffett Field, CA, USA, by the AVIRIS spectro-imager in 1997 and has been used in several studies, \eg \cite{Christophe2005,Akgun2005,Dobigeon2009}. The scene was captured in 224 bands, representing the spectrum from \unit{400}{\nano\meter} to \unit{2500}{\nano\meter}. 
Noisy bands have been removed, leaving 188 bands for the evaluation. As in \cite{Dobigeon2008,Dobigeon2009}, a subset of $50\times50$ pixels size is chosen, where each pixel represents an area of $\unit{20}{\meter} \times \unit{20}{\meter}$.
The subset is depicted in \figref{fig::res::moffett::real}, showing a coastal scenery.

The results in \figref{fig::res::moffett::emam} reveal three different endmembers: grass, soil, and water, which is in line with the results found in \cite{Dobigeon2009}. \ac{VCA} and \ac{BNU} provide results of similar accuracy, with \ac{BNU} estimating three endmembers in each of the five Monte Carlo runs. 
The abundances clearly show the presence of water, soil and grass, as typical for a coastal scenery.

\subsubsection{Cuprite}
\label{res::real::cuprite}
The Cuprite scene has been extensively investigated in \ac{HSI} research. Like the Moffett field, this scene has been captured by the AVIRIS imager in 1997 and shows a spatial resolution of $\unit{20}{\meter}$.
The scene covers the Cuprite mining area in Nevada, USA, and, thus, mainly contains different minerals. 
A subset is chosen as depicted in \figref{fig::res::cuprite::real}, showing mainly three different materials, as detailed in \cite{Dobigeon2009}. 
The endmembers in this scene show strong correlations, in contrast to those of the Moffett field, which renders the extraction of the endmembers challenging.
After convergence to the stationary distribution, the Gibbs sampler in \ac{BNU} creates samples with four and five endmembers.
A sample with 5 endmembers is shown in \figref{fig::res::cuprite::emam}, where the spectral range is limited between $\unit{2}{\micro\meter}$ and $\unit{2.4}{\micro\meter}$ for detailed analysis.
\ac{BNU} provides a more accurate reconstruction of the endmembers than \ac{VCA}, especially for Kaolinite.
The additionally extracted endmembers are similar to Kaolinite and Alunite. Merging similar endmembers would yield results comparable with those in \cite{Dobigeon2009}.

\section{Discussion and Outlook}
\label{sec::discussion}
As demonstrated in the experiments, \ac{BNU} is able to accurately extract the endmembers and estimate the abundances, providing results comparable with state-of-the-art algorithms.
Additionally, \ac{BNU} infers the number of endmembers from the data automatically and, thus, does not require any prior knowledge about the scene of interest. 
\ac{SPICE} is also able to estimate this information jointly, but often results in less accurate estimates in comparison with \ac{BNU}, especially in the presence of strong noise. 

\begin{figure}
  \centering
  \includegraphics[scale=.67]{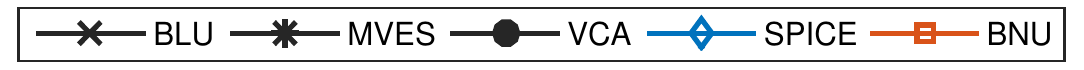}\\
  \includegraphics{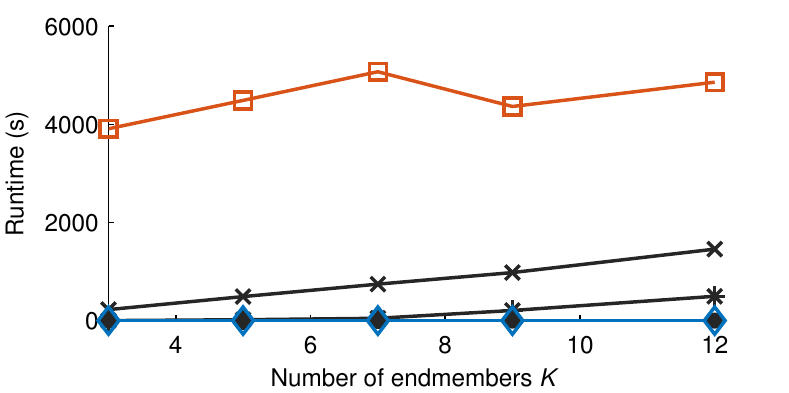}
  \caption{Runtime comparison of the implementations of the algorithms. Due to parallel tempering (5 Monte Carlo chains), \ac{BNU} is the slowest approach, significantly slower than the other approaches.}
  \label{fig::dis::dim::k::runtime}
\end{figure}

One drawback of \ac{BNU} is, as common for algorithms based on Bayesian inference, the runtime of the algorithm.
As an example, in \figref{fig::dis::dim::k::runtime}, we show the runtime of the unoptimized implementations of the algorithms for the simulation of different numbers of endmembers.
Note that we ran five Markov chains for \ac{PT} on a single core architecture for \ac{BNU}. 
Thus, to obtain the results for a single chain, \ie if \ac{PT} was not used, the runtime scales down to a fifth of the shown values. 
For better comparison, we set the number of samples to $1,000$, which is sufficient in this scenario. 
\ac{BLU} usually requires far less samples and is, therefore, much faster, thanks to a better initialization of the underlying Markov chain by running another endmember extraction algorithm first.

In order to reduce the runtime, a variational approach of the \ac{IBP} has been proposed in \cite{Doshi-Velez2009}.
The authors in \cite{Doshi-Velez2009} found that, depending on the application, the approximations introduced by the variational algorithm lead to inaccurate and slow inference such that in some cases, Gibbs sampling is not only more accurate, but also faster.
Further, it is reported that the variational approach succeeded only when the number of variables was sufficiently large.
As the \ac{IBP} is related to the Beta-process \cite{Thibaux2007}, variational approaches from this field may yield better solutions, \eg \cite{Carin2011}. 

Relying on Bayesian inference, \ac{BNU} tries to explain the observed data and expresses its belief over the unknowns by means of the posterior.
Noisy observations can have a two-fold effect on the inference: either the variance of the noise is increased or the number of endmembers is overestimated, where the noise is basically absorbed into the additional endmembers.
In particular, this effect can be observed if the observations suffer from noise which cannot be explained by the model, as shown in the experiments in \secref{res::sim::illum} and \secref{res::real::cuprite}. 

Consequently, in the case of strong noise, there often exist several, almost equal probable explanations for the observations. 
The approximate \ac{MAP} estimator we used, however, considers only one explanation as the estimate consists of the most probable sample only (\secref{sec::inf::sampling}). 
Thus, the \ac{MAP} estimate contains limited information about the posterior.
Better generalization capabilities are provided by estimators that take the shape of the posterior into account, \eg the \ac{MMSE} estimator. 
The \ac{MMSE} estimator, however, cannot be used with the \ac{IBP} to infer the endmembers due to the varying dimensionality of the samples.
Hence, the problem remains to find better estimators that can be utilized in \ac{BNU}.

Though the prior used in \ac{BNU} works reasonably well, it has the disadvantage that the hyperparameter $\hyperWgamma$ has to be set \apriori.
Observing the prior for the feature weights $\Wmat$, it becomes clear that the prior favors similar bands (and hence similar endmembers), where $\hyperWgamma$ controls the tolerated variance. If we set this parameter too high (low variance), then newly introduced endmembers are likely to be rejected or merged with existing endmembers. 
Hence, an important future direction is the development of a suitable prior for $\Wmat$ with known normalization, such that the hyperparameters can be sampled efficiently, and thus, be learned from the observations.
Alternatively, if a database of many data sets with known parameters is given, meta-learning \cite{Vilalta2002} is an option. 
Meta-learning measures the similarity between the data sets and the observed data, aiming at exploiting the parameters stored in the database for the new data.

\section{Conclusion}
\label{sec::conclusion}
We have presented a Bayesian nonparametric framework for hyperspectral unmixing that is able to jointly estimate the endmembers, their fractional abundances, and, in contrast to most existing algorithms, the number of endmembers.
This framework borrows from probabilistic feature learning concepts by modeling the endmember by means of an \ac{IBP}, allowing to infer the number of endmembers from the observed data.
In contrast to most previous work, the number of endmembers, thus, does need to be set \apriori. Inference in this hierarchical Bayesian model is accomplished by means of Gibbs sampling. 
Due to the high flexibility of the model, the sampler might get trapped in a mode of the posterior. We propose to solve this problem by making use of parallel tempering. 
Experimental results on simulated and real data demonstrate the performance of this approach, which is comparable to state-of-the-art algorithms, while additionally estimating the number of endmembers.
Future directions can include variational inference to speed up inference and the investigation of different priors for the endmembers.

\bibliographystyle{plain}
\bibliography{hsi}

\end{document}